\pdfoutput=1

\documentclass[11pt]{article}

\usepackage{EMNLP2022}

\usepackage{times}
\usepackage{latexsym}

\usepackage[T1]{fontenc}

\usepackage[utf8]{inputenc}

\usepackage{microtype}

\usepackage{inconsolata}

\usepackage{mathtools}
\usepackage{times}
\usepackage{latexsym}
\usepackage{epsfig}
\usepackage{graphicx}
\usepackage{amsmath}
\usepackage{amssymb}
\usepackage{multirow}
\usepackage{multicol}
\usepackage{adjustbox}
\usepackage{booktabs} 
\usepackage{makecell}
\usepackage{pifont}
\usepackage[T1]{fontenc}
\usepackage{mathabx}
\usepackage{subcaption}
\usepackage{colortbl}
\usepackage{soul}
\usepackage{array}
    \newcolumntype{P}[1]{>{\centering\arraybackslash}p{#1}}
    \newcolumntype{M}[1]{>{\centering\arraybackslash}m{#1}}
 
\usepackage{enumitem}
\usepackage{hyperref}

\def\CircleArrowright{\ensuremath{%
  \rotatebox[origin=c]{310}{$\circlearrowright$}}}

\DeclareMathOperator*{\argmax}{arg\,max}
\DeclareMathOperator*{\amax}{max}

\title{ULN: Towards Underspecified Vision-and-Language Navigation}

\author{Weixi Feng \,\, Tsu-Jui Fu \,\, Yujie Lu \,\, William Yang Wang \\ \\
    UC Santa Barbara \\
    \texttt{\{weixifeng, tsu-juifu, yujielu, william\}@cs.ucsb.edu} \\ }

\begin{document}
\maketitle

\begin{abstract}
Vision-and-Language Navigation (VLN) is a task to guide an embodied agent moving to a target position using language instructions. Despite the significant performance improvement, the wide use of fine-grained instructions fails to characterize more practical linguistic variations in reality. To fill in this gap, we introduce a new setting, namely \textbf{U}nderspecified vision-and-\textbf{L}anguage \textbf{N}avigation (ULN), and associated evaluation datasets. ULN evaluates agents using multi-level underspecified instructions instead of purely fine-grained or coarse-grained, which is a more realistic and general setting. As a primary step toward ULN, we propose a VLN framework that consists of a classification module, a navigation agent, and an Exploitation-to-Exploration (E2E) module. Specifically, we propose to learn Granularity Specific Sub-networks (GSS) for the agent to ground multi-level instructions with minimal additional parameters. Then, our E2E module estimates grounding uncertainty and conducts multi-step lookahead exploration to improve the success rate further. Experimental results show that existing VLN models are still brittle to multi-level language underspecification. Our framework is more robust and outperforms the baselines on ULN by ~10\% relative success rate across all levels. \footnote{Our code and data are available at \url{https://github.com/weixi-feng/ULN}.}
\end{abstract}

\section{Introduction}
Vision-and-Language Navigation (VLN) allows a human user to command or instruct an embodied agent to reach target locations using verbal instructions. For this application to step out of curated datasets in real-world settings, the agents must generalize to many linguistic variations of human instructions. Despite significant progress in VLN datasets~\cite{anderson2018vision, chen2019touchdown, ku2020room, shridhar2020alfred} and agent design~\cite{fried2018speaker, li2021improving, min2021film}, it remains a question whether existing models are generalized and robust enough to deal with all kinds of language variations.

\begin{figure}[t]
    \centering
    \includegraphics[width=\linewidth]{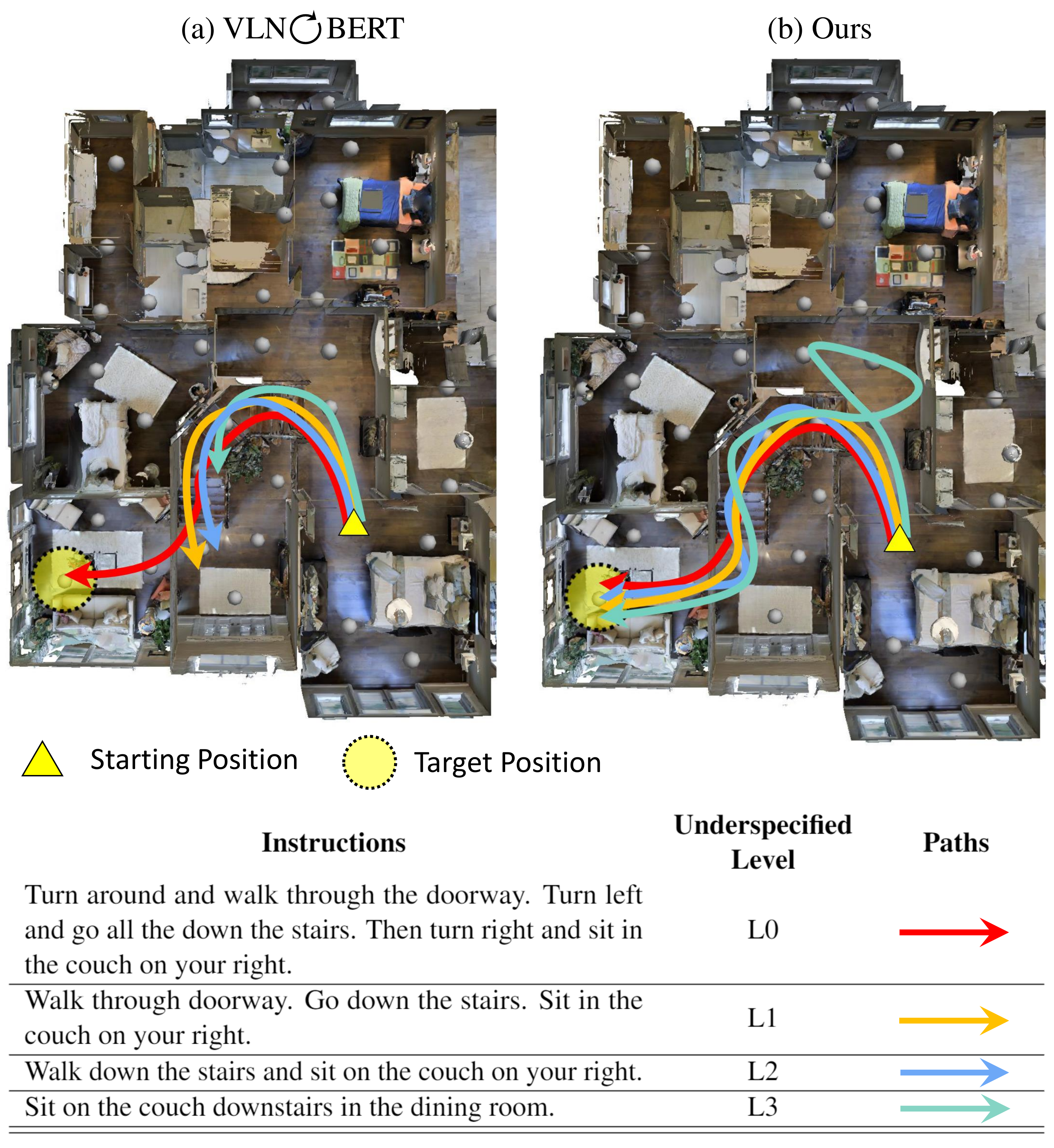}
    \caption{Navigation results of a baseline (left) and our VLN framework (right) with multi-level underspecified instructions~($L_0$-$L_3$). Trajectories are curved for demonstration. Note that the baseline stops early and fails to reach the target position with $L_1$-$L_3$. Our agent manages to reach the goal across all levels.}
    \label{fig:teaser}
\end{figure}

For the language input in an indoor environment, some datasets focus on long and detailed instructions with the route description at every step to achieve fine-grained language grounding~\cite{anderson2018vision, ku2020room} or long-horizon navigation~\cite{jain2019stay, zhu2020babywalk}. For instance, from Room-to-Room~(R2R)~\cite{anderson2018vision}, to Room-Across-Room~(RxR)~\cite{ku2020room}, the average instruction length increases from 29 to 129 words. Other datasets have coarse-grained instructions like REVERIE \cite{qi2020reverie} or SOON \cite{zhu2021soon}. Agents are trained and evaluated on a single granularity or one type of expression.   

In contrast, we propose to evaluate VLN agents on multi-level granularity to better understand the behavior of embodied agents with respect to language variations. Our motivation is that users are inclined to give shorter instructions instead of detailed route descriptions because 1) users are not omniscient observers who follow the route and describe it step by step for the agent; 2) shorter instructions are more practical, reproducible, and efficient from a user's perspective. 3) users tend to underspecify commands in familiar environments like personal households. Therefore, we propose a new setting, namely \textbf{U}nderspecified vision-and-\textbf{L}anguage \textbf{N}avigation~(ULN) and associated evaluation datasets on top of R2R, namely R2R-ULN to address these issues. R2R-ULN contains underspecified instructions where route descriptions are successively removed from the original instructions. Each long R2R instruction corresponds to three shortened and rephrased instructions belonging to different levels, which preserves partial alignment but also introduces variances.

As shown in Fig.~\ref{fig:teaser}, the goal of ULN is to facilitate the development of a generalized VLN design that achieves balanced performance across all granularity levels. As a primary step toward ULN, we propose a modular VLN framework that consists of an instruction classification module, a navigational agent, and an Exploitation-to-Exploration (E2E) module. The classification module first classifies the input instruction as high-level or low-level in granularity so that our agent can encode these two types accordingly. As for the agent, we propose to learn Granularity Specific Sub-networks (GSS) to handle both levels with minimally additional parameters. A sub-network, e.g., the text encoder, is trained for each level while other parameters are shared. Finally, the E2E module estimates the step-wise language grounding uncertainty and conducts multi-step lookahead exploration to rectify wrong decisions that originated from underspecified language.

Our VLN framework is model-agnostic and can be applied to many previous agents that follow a ``encode-then-fuse'' mechanism for language and visual inputs. We establish our framework based on two state-of-the-art (SOTA) VLN agents to demonstrate its effectiveness. We conduct extensive experiments to analyze the generalization of existing agents and our framework in ULN and the original datasets with fine-grained instructions. Our contribution is three-fold:

\begin{itemize}
    \item We propose a novel setting named Underspecified vision-and-Language Navigation (ULN) to account for multi-level language variations for instructions. We collect a large-scale evaluation dataset R2R-ULN which consists of $9$k validation and $4$k testing instructions.

    \item We propose a VLN framework that consists of Granularity Specific Sub-networks (GSS) and an E2E module for navigation agents to handle both low-level and high-level instructions. 

    \item Experiments show that achieving consistent performance across multi-level underspecification can be much more challenging to existing VLN agents. Furthermore, our VLN framework can improve the success rate by ~10\% relatively over the baselines and mitigate the performance gap across all levels. 
\end{itemize}

\section{Related Work}

\paragraph{Language Variations for Multimodal Learning}
Natural language input has been an essential component of modern multimodal learning tasks to combine with other modalities such as vision \cite{antol2015vqa, johnson2017clevr}, speech \cite{alayrac2020self} or gestures \cite{chen2021yourefit}. The effect of language variations has been studied in many vision-and-language (V\&L) tasks \cite{bisk2016natural, agrawal2018don, cirik2018visual, zhu2020towards, lin2021adversarial}. For instance, referring expression datasets \cite{kazemzadeh2014referitgame, yu2016modeling, mao2016generation} contain multiple expressions for the same referring object. Ref-Adv \cite{akula2020words} studies the robustness of referring expression models by switching word orders. In Visual Question Answering (VQA), \citeauthor{shah2019cycle} (\citeyear{shah2019cycle}) discovers that VQA models are brittle to rephrased questions with the same meaning. 
As for VLN, we characterize the linguistic and compositional variations in rephrasing and dropping sub-instructions from a full instruction with complete route descriptions. We also define three different levels to formalize underspecification for navigational instructions.

\paragraph{VLN Datasets} VLN has gained much attention \cite{gu2022vision} with emergence of various simulation environments and datasets \cite{chang2017matterport3d, kolve2017ai2, jain2019stay, nguyen2019help, koh2021pathdreamer}. R2R \cite{anderson2018evaluation} and RxR \cite{ku2020room} provide fine-grained instructions which guide the agent in a step-wise manner. FG-R2R \cite{hong2020sub} and Landmark-RxR \cite{he2021landmark} segments the instructions into action units and explicitly ground sub-instructions on visual observation. In contrast, REVERIE \cite{qi2020reverie}, and SOON \cite{zhu2021soon} proposes to use referring expression with no guidance on intermediate steps that lead to the final destination. Compared to these datasets, ULN aims to build an agent that can generalize to multi-level granularity after training once, which is more practical for real-world applications.

\paragraph{Embodied Navigation Agents} Learning to ground instructions on visual observations is one major problem for an agent to generalize to an unseen environment \cite{wang2019reinforced, deng2020evolving, fu2020counterfactual}. Previous studies demonstrate significant improvement by data augmentations \cite{fried2018speaker, tan2019learning, zhu2021multimodal, fang2022stemm, li2022envedit}, designing pre-training tasks \cite{hao2020towards, chen2021history, qiao2022hop} and decoding algorithms \cite{ma2019selfmonitoring, ke2019tactical, ma2019regretful, chen2022think}. For exploration-based methods, FAST \cite{ke2019tactical} proposes a searching algorithm that allows the agent to backtrack to the most promising trajectory. SSM \cite{wang2021structured} memorizes local and global action spaces and estimates multiple scores for candidate nodes in the frontier of trajectories. Compared to E2E, Active VLN \cite{wang2020active} is the most relevant work where they learn an additional policy for multi-step exploration. However, they define the reward function based on distances to target locations, while our uncertainty estimation is based on step-wise grounding mismatch. Our E2E module is also more efficient that has fewer parameters and low training complexity. 

\section{Underspecification in VLN}
\label{sec:dataset}

\begin{table}[t]
\begin{center}
  \resizebox{\linewidth}{!}{
  \begin{tabular}{l|l}
    \toprule
    \multicolumn{1}{c|}{\textbf{Level}} & \multicolumn{1}{c}{\textbf{Instructions}} \\
    \midrule \midrule
    $L_{0}$ & \makecell[l]{\textcolor{red}{Turn around} and \textcolor{blue}{go down the stairs. At the bottom} \\ \textcolor{red}{turn slightly}
                        \textcolor{red}{right} and enter the room with the TV \\ on the wall and a green table. 
                        Walk \textcolor{red}{to the right} \\ past the TV. Stop at the door \textcolor{red}{to the right} facing \\
                        into the bathroom. (\textit{from R2R})} \\
    \midrule
    $L_{1}$ & \makecell[l]{Take the stairs to the bottom and enter the room \\ with \textcolor{red}{the TV on} 
                        \textcolor{red}{the wall} and a green table. \\Walk past the TV. Stop at the door to 
                        \textcolor{red}{facing into} \\ the bathroom. (\textit{Redundancy Removed})}\\
    \midrule
    $L_{2}$ & \makecell[l]{\textcolor{red}{Take the stairs to the bottom and} enter the room \\ a green table. 
                        \textcolor{red}{Walk past the TV}. Stop at the\\ bathroom door. (\textit{Partial Route} 
                        \textit{Description}) }\\
    \midrule
    $L_{3}$ & \makecell[l]{Go to the door of the bathroom next to the room \\ with a green table. (\textit{No Route Description})}\\
    \midrule
    \midrule
  \end{tabular}
   }
\end{center}
\caption{Instruction examples from the R2R-ULN validation set. We mark removed words in red and rephrased words in blue in the next level.}
\label{tab:uln_showcase}
\end{table}

Our dataset construction is three-fold: We first obtain underspecified instructions by asking workers to simplify and rephrase the R2R instructions. Then, we validate that the goals are still reachable with underspecified instructions. Finally, we verify that instructions from R2R-ULN are preferred to R2R ones from a user's perspective, which proves the necessity of the ULN setting. We briefly describe definitions and our ULN dataset in this section with more details in Appendix \ref{app:dataset}.

\subsection{Instruction Simplification}
\label{subsec:data_collect}
We formalize the instruction collection as a sentence simplification task and ask human annotators to remove details from the instructions progressively. Denoting the original R2R instructions as \textbf{Level 0}~($L_0$), annotators rewrite each $L_0$ into three different levels of underspecified instructions. We discover that some components in $L_0$ can be redundant or rarely used in indoor environments, such as ``turn 45 degrees''. Therefore, to obtain \textbf{Level 1} ($L_1$) from each $L_0$ instruction, annotators rewrite $L_0$ by removing any redundant part but keep most of the route description unchanged. Redundant components include but are not limited to repetition, excessive details, and directional phrases (See Table \ref{tab:uln_showcase}). As for \textbf{Level 2} ($L_2$), annotators remove one or two sub-instructions from $L_1$, creating a scenario where the users omit some details in commonplaces. We collect \textbf{Level 3} ($L_3$) instructions by giving destination information such as region label and floor level and ask annotators to write one sentence directly referring to the object or location of the destination point. 

\begin{table}[t]
  \begin{center}
  \resizebox{0.95\linewidth}{!}{
  \begin{tabular}{c|c|c|c|c}
    \hline \hline
    \multicolumn{1}{c|}{\multirow{3}{*}{Level}} & \multicolumn{4}{c}{R2R-ULN Val-Unseen} \\
    \cline{2-5} &
    \multicolumn{2}{c|}{Instr. Following} & \multicolumn{2}{c}{Instr. Preference} \\
    \cline{2-5} &
    \multicolumn{1}{c|}{SR$\uparrow$} & \multicolumn{1}{c|}{SPL$\uparrow$} &
    \multicolumn{1}{c|}{Practicality} & \multicolumn{1}{c}{Efficiency}\\
    \hline \hline
    $L_0$ & 86 & 72 & -    & - \\
    $L_1$ & 82 & 68 & 55\% & 57\% \\
    $L_2$ & 82 & 65 & 63\% & 59\% \\
    $L_3$ & 75 & 58 & 68\% & 66\%  \\
    \hline \hline
  \end{tabular}
  }
\end{center}
\caption{Human performance on R2R-ULN validation unseen in terms of Success Rate~(SR) and SR weighted by Path Length~(SPL), and human preference assessment results. The percentage denotes the ratio of participants selecting $L_i$ over $L_0$.}
\label{tab:human_verify}
\end{table}

\subsection{Instruction Verification}\label{subsec:data_verification}
To ensure that the underspecified instructions provide a feasible performance upper bound for VLN agents, we have another group of annotators navigate in an interactive interface from R2R \cite{anderson2018vision}. As is shown in Table~\ref{tab:human_verify}, annotators achieve a slightly degraded but promising success rate (SR) with $L_3$. SPL is a metric that normalizes SR over the path length. Therefore, the trade-off for maintaining high SR is to have more exploration steps, resulting in a much lower SPL value. We also verify that $L_i, i\in\{1,2,3\}$ are more practical and efficient choices than $L_0$. Table \ref{tab:human_verify} shows that people prefer underspecified instructions over full instructions in both aspects, with an increasing trend as $i$ increases to 3.

\section{Method}
\label{sec:method}

\subsection{Overview}
In this section, we present our VLN framework for handling multi-level underspecified language inputs, which mainly consists of three modules (see Figure \ref{fig:framework}). Given a natural language instruction in a sequence of tokens, $\mathcal{W}=(w_1,\ldots w_n)$ ,the classification module first categorize language input as low-level ($L_0, L_1, L_2$) or high-level ($L_3$) instructions. To handle these two types accordingly, GSS learns a sub-network, e.g., the text encoder, for each type while the other parameters are shared. At each step $t$, we denote the visual observation $\mathcal{O}_t=([v_1;a_1],\ldots , [v_N, a_N])$ with visual feature $v_i$ and angle feature $a_i$ of $i$-th view among all $N$ views. The history contains a concatenation of all observations before $t$m $H_t=(\mathcal{O}_1,\ldots ,\mathcal{Q}_{t-1})$. Given $\mathcal{W}_t, \mathcal{H}_t, \mathcal{O}_t$, the GSS-based agent predicts a an action $a_t$ by choosing a navigable viewpoint from $\mathcal{O}_t$. To overcome the reference misalignment issue, the E2E module predicts a sequence of uncertainty score $\mathcal{S} = (s_1, ..., s_T)$ and conducts multi-step exploration to collect future visual information.

\subsection{Instruction Classification}
\label{subsec:module_class}
VLN agents can operate in two different modes, fidelity-oriented or goal-oriented, depending on reward functions \cite{jain2019stay} or text inputs \cite{zhu2022diagnosing}. Agents trained on low-level granularity encounter performance degradation when applied to high-level ones, and vice versa. As is shown in Figure \ref{fig:framework}, we propose first to classify the text inputs into two granularities and then encode them independently in downstream modules. Our classification module contains an embedding layer, average pooling, and a fully-connected layer to output binary class predictions.

\subsection{Navigation Agent}
\label{subsec:module_agent}

\begin{figure}[t]
    \centering
    \includegraphics[width=\linewidth]{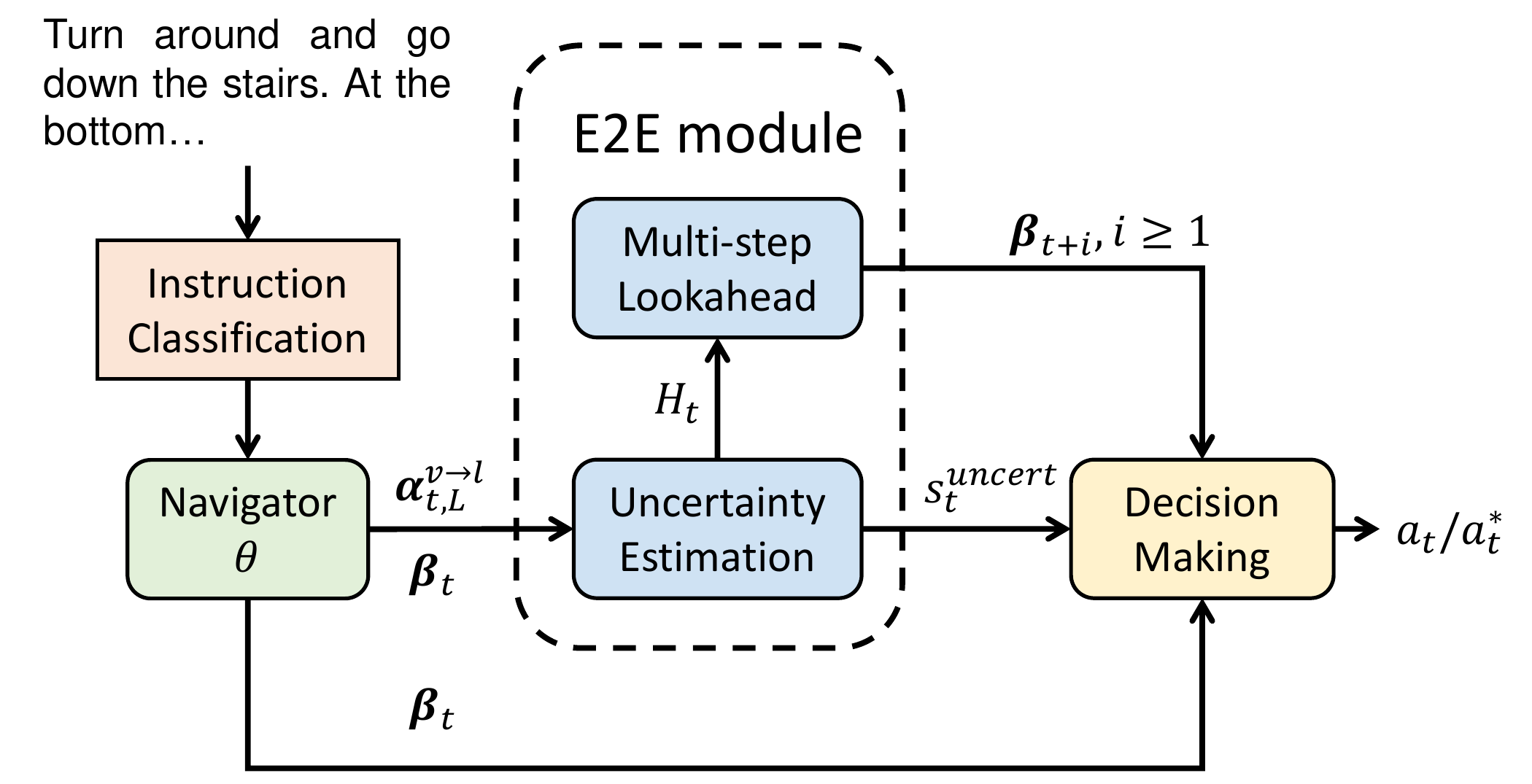}
    \caption{Our VLN framework with classification module, navigation agent, and E2E module.}
    \label{fig:framework}
\end{figure}

\paragraph{Base Agent} We summarize the high-level framework of many transformer-based agents \cite{hao2020towards, guhur2021airbert, moudgil2021soat} paramterized as $\theta$ as shown in Figure \ref{fig:framework}. Given the history $\mathcal{H}_t$, text $\mathcal{X}$, visual observation $\mathcal{O}_t$, the agent first encodes each modality input with encoders $f_{\text{hist}}, f_{\text{text}}, f_{\text{img}}$:
\begin{equation}
\begin{array}{ll}
    X = f_{\text{text}}(\mathcal{W}), & H_t = f_{\text{hist}}(\mathcal{H}_t),\\ O_t = f_{\text{img}}(\mathcal{O}_t)
\end{array}
\end{equation}

HAMT \cite{chen2021history} applies ViT \cite{dosovitskiy2020image} and a Panoramic Transformer to hierarchically encode $\mathcal{H}_t$ as a sequence of embeddings $H_{t}=(h_1,\ldots, h_{t-1})$ while VLN$\protect\CircleArrowright$BERT \cite{hong2021vln} encodes $\mathcal{H}_t$ as a state vector $H_{t} = h_{t}$. The embedding from each modality is then fed into a $L$-layer cross-modal transformer $f_{\text{cm}}$, and passed through a cross-attention first in each layer $l$:
\begin{equation}
    \boldsymbol{\alpha}_{t,l}^{\text{v}\rightarrow \text{t}} = \frac{([H_{t,l};O_{t,l}]W_{l}^{\text{query}})(X_{t,l}W_{l}^{\text{key}})^T}{\sqrt{d_h}}
\end{equation}
where $\alpha_{t,l}^{\text{v}\rightarrow \text{t}}$ denotes the attention weights of history-visual concatenation on the language embeddings, $d_h$ is the hidden dimension. We omit the attention head index for simplicity. For VLN$\protect\CircleArrowright$BERT, it concatenated state $H_t$ with $X_t$ instead. The prediction of $a_t$ relies on either a two-layer FC network $f_{\text{action}}$, or the summation of attention weights of $H_t$ on $O_t$ over all heads:
\begin{align}
    \boldsymbol{{\beta}}_{t}^{\text{HAMT}} &= f_{\text{action}} (O'_{t}\odot x'_{t, 1}) \\
    \boldsymbol{{\beta}}_{t}^{\text{VLN\protect\CircleArrowright BERT}} &= \sum_{\text{head}} \frac{(H_{t,L}W_{L}^{Q})(O_{t,L}W_{L}^{K})^T}{\sqrt{d_h}} \\
    a_t &= \argmax_{c} (\beta_{t, c})
\end{align}
where $O'_{t}, X'_{t}$ are the observation and language tokens output from $f_{\text{cm}}$.

\begin{figure}[t]
    \centering
    \includegraphics[width=\linewidth]{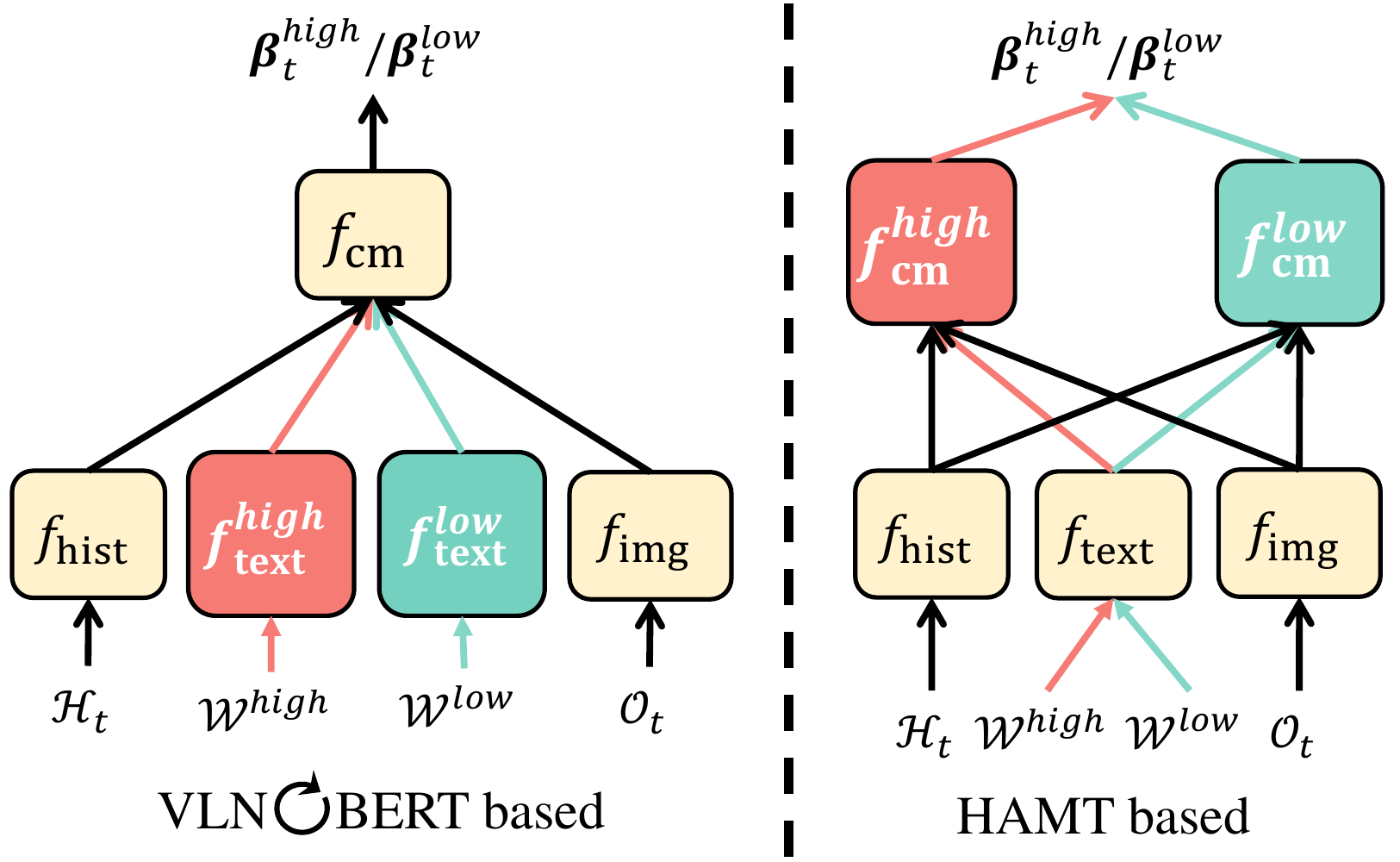}
    \caption{Granularity Specific Sub-networks of two different agents. We discover that the critical sub-network for VLN$\protect\CircleArrowright$BERT is $f_{\text{text}}$, while $f_{\text{cm}}$ for HAMT. }
    \label{fig:gss}
\end{figure}

\paragraph{Granularity Specific Sub-network} Training an ensemble of agents or one agent with a mixture of levels can be inefficient or sub-optimal for ULN. Instead, we find a sub-network that influences the agent's navigation mode, as shown in Figure \ref{fig:gss}. We identify such sub-network by the following steps:
\begin{enumerate}
    \item Train an agent $\theta_{\text{l}}$ with full instructions $L_0$ and a separate agent $\theta_{\text{h}}$ with the last sentence of $L_0$. Denote $\theta_{\text{l}}$ performance on $L_3$ as metric value $m_{\text{l}\rightarrow \text{h}}$.
    \item For each of the sub-network, $f_{\text{text}}$, $f_{\text{img}}$, $f_{\text{hist}}$, and $f_{\text{cm}}$ in $\theta_{\text{h}}$, load its weights to $\theta_{\text{l}}$ and denote the performance on $L_3$ as $m^{x}_{\text{l}\rightarrow \text{h}}$ where $x\in \{\text{text, img, hist, cm}\}$ indicates the sub-network replaced.
    \item Find the sub-network with the maximum gain in metric value on $L_3$ after replacement, i.e. $x^{*} = \argmax_{x} (m^{x}_{\text{l}\rightarrow \text{h}} - m_{\text{l}\rightarrow \text{h}})$.
\end{enumerate}

After identifying the critical sub-network $f_{x^{*}}$, we train a new $f_{x^{*}}$ from scratch with the rest of the model parameters loaded from $\theta_{l}$ and kept frozen. 

\subsection{Exploitation to Exploration}\label{subsec:e2e}
Multi-level inputs introduces \textbf{Temporal Reference Misalignment~(TRM)}. As the agent gradually shifts its attention to sub-instructions, it lacks a mechanism to ensure the attended text segments align with the visual observation transition. Consequently, after several steps, agents cannot correctly ground sub-instructions to visual features. To mitigate this issue, we propose an E2E module to estimate step-wise uncertainty and perform multi-step lookahead to skip the dilemma.

\paragraph{Uncertainty Estimation} We evaluate the decision uncertainty at each step based on the \textbf{attention score distribution}. TRM changes the distribution of $\boldsymbol{\alpha}_{t,L}^{\text{v}\rightarrow \text{t}}, \boldsymbol{\beta}_{t}$ and makes them different from the distribution when full instructions are given. Therefore, the joint distribution of $\boldsymbol{\alpha}_{t,L}^{\text{v}\rightarrow \text{t}}, \boldsymbol{\beta}_{t}$ implies the degree of grounding uncertainty. We simply input the concatenation $\boldsymbol{\alpha}_{t,L}^{\text{v}\rightarrow \text{t}}, \boldsymbol{\beta}_{t}$ to a two-layer MLP $f_{\text{uncert}}$ to learn the uncertainty score:
\begin{align}
    s^{\text{uncert}}_{t} = f_{\text{uncert}}([\boldsymbol{\alpha}_{t,L}^{\text{v}\rightarrow \text{t}};\boldsymbol{\beta}_{t}]).
\end{align}
$s^{\text{uncert}}_{t} \in [0,1]$ indicates whether the agent is confident for the decision. If $s^{\text{uncert}}$ is greater than a threshold, the agent first explores the environment before the next step decisions.

\begin{table*}[t]
  \begin{center}
  \resizebox{\textwidth}{!}{
  \begin{tabular}{l c c cccc c cccc}
    \hline \hline
    \multicolumn{1}{c}{\multirow{3}{*}{Models}} & \multicolumn{1}{c}{\multirow{3}{*}{Training Set}} & & \multicolumn{4}{c}{\multirow{2}{*}{R2R Val-Unseen}} & & \multicolumn{4}{c}{R2R-ULN Val-Unseen}\\
    & & & & & & & & \multicolumn{4}{c}{Level 3 ($L_3$)}\\
    \cline{4-7} \cline{9-12} & & & 
    \multicolumn{1}{c}{TL} & \multicolumn{1}{c}{NE$\downarrow$} & \multicolumn{1}{c}{SR$\uparrow$} & \multicolumn{1}{c}{SPL$\uparrow$} & & \multicolumn{1}{c}{TL} & \multicolumn{1}{c}{NE$\downarrow$} & \multicolumn{1}{c}{SR$\uparrow$} & \multicolumn{1}{c}{SPL$\uparrow$} \\
    \hline
    \texttt{1} VLN$\protect\CircleArrowright$BERT & R2R & & 12.09 & 4.10 & \underline{61.4} & \underline{55.6} && 13.20 & 7.41 & 32.7 & 29.1 \\
    \texttt{2} VLN$\protect\CircleArrowright$BERT  & R2R-last & & 12.34 & 5.01 & 53.8 & 47.7 && 13.31 & \underline{6.96} & \underline{36.5} & \underline{32.0} \\
    \texttt{3} VLN$\protect\CircleArrowright$BERT & R2R+R2R-last & & 11.78 & 4.28& 59.2 & 53.5 && 12.81 & 7.12 & 35.7 & 31.2 \\
    \texttt{4} Ours (w/o E2E) & R2R+R2R-last & & 12.09 & \textbf{4.08} & \textbf{61.6} & \textbf{55.8} && 13.30 & \textbf{6.91} & \textbf{37.8} & \textbf{33.6} \\
    \hline
    \texttt{5} HAMT & R2R && 11.46 & \textbf{3.62} & \textbf{66.2} & \textbf{61.5} && 13.36 & 7.18 & 35.1 & 31.1 \\
    \texttt{6} HAMT  & R2R-last && 11.57 & 4.55 & 57.1 & 52.4 && 13.51 & \underline{6.84} & \underline{37.6} & \underline{33.9}\\
    \texttt{7} HAMT & R2R+R2R-last && 11.12 & 3.90 & 63.8 & 30.1 && 13.20 & 7.13 & 36.3 & 32.4 \\
    \texttt{8} Ours (w/o E2E) & R2R, R2R-last && 11.54 & \underline{3.71} & \underline{65.6} & \underline{60.7} && 13.26 & \textbf{6.75} & \textbf{38.8} & \textbf{34.9} \\
    \hline \hline
  \end{tabular}
  }
\end{center}
\caption{Comparison of two baselines and our VLN framework with a classification module and GSS. We bold the best values and underline the second-best values.}
\label{tab:dual_encoder_result}
\end{table*}

\paragraph{Multi-Step Lookahead} When score $s^{\text{uncert}}_{t}$ indicates an uncertain decision, our system calculates the likelihood of success by exploration and re-evaluates the action logits. Specifically, the explorer moves $K$ steps forward for each of the top $C$ candidate actions at step $t$. Since unnormalized logits incorporate alignment between actions and instructions, we adopt the attention weights on visual candidates, i.e., $\boldsymbol{\beta}_{t}$. The new action probability estimation accounts for a weighted sum of the future logits sequence with a hyperparameter $\gamma$:
\begin{align}
\label{eq:lookahead}
    a_{t}^{*} = \argmax_{c} \big[\beta_{t, c} + \sum_{i=1}^{K} \gamma^{i} \amax_{c'} (\beta_{t+i, c'})\big].
\end{align}
$\beta_{t, c}$ is the logit value for candidate $c$, and $\boldsymbol{\beta}_{t+i}$ are the logits after executing greedy action at step $t+i-1$. For parameter efficiency, we utilize the trained agent as the explorer. Our lookahead heuristic differentiates from Active VLN \cite{wang2020active} as we explicitly quantify the misalignment by uncertainty estimation, while the latter depends on goal-based reward functions with implicit supervision. Our E2E module is also more efficient and stable as we spare the need to train a separate policy for exploration.

\paragraph{State Freeze} The history encoding $h_t$ serves as an important query to attend to on both visual and language domains. Due to the underspecified language input, the history-attended instruction advances the transition in the visual scenarios. Therefore, to calculate $\beta_{t+1, c'}$ in Equation~\ref{eq:lookahead}, we shall utilize $h_{t-1}$ instead of $h_{t}$ to maintain the attention on the pending sub-instruction until alignment recovers.

\section{Experiments}\label{sec:exp}

\begin{table*}[t]
  \begin{center}
  \resizebox{\textwidth}{!}{
  \begin{tabular}{ll c cccc c cccc}
    \hline \hline
    &\multicolumn{1}{c}{\multirow{2}{*}{Methods}} && \multicolumn{4}{c}{R2R-ULN Val-Seen} && \multicolumn{4}{c}{R2R-ULN Val-Unseen} \\
    \cline{4-7} \cline{9-12} &&
    &\multicolumn{1}{c}{TL} & \multicolumn{1}{c}{NE$\downarrow$} & \multicolumn{1}{c}{SR$\uparrow$} & \multicolumn{1}{c}{SPL$\uparrow$} &&
    \multicolumn{1}{c}{TL} & \multicolumn{1}{c}{NE$\downarrow$} & \multicolumn{1}{c}{SR$\uparrow$} & \multicolumn{1}{c}{SPL$\uparrow$} \\
    \hline \hline
    \texttt{0} & Human
        && -     & - & - &  -   &&  14.97   & 2.94 & 77.4 & 61.7 \\
    \hline
    &\textit{Greedy-Decoding Agents} \\
    \texttt{1} & Speaker-Follower~\cite{fried2018speaker}
        && 12.26     & 6.17 & 43.4 &  39.4   &&  14.86   & 8.43 & 22.0 & 17.4     \\
    \texttt{2} & SMNA~\cite{ma2019selfmonitoring}
        && 12.07     & 5.95 & 46.5 &  41.0   &&  15.42   & 7.98 & 23.1 & 16.3  \\
    \texttt{3} & EnvDrop~\cite{tan2019learning}
        && 9.14      & 6.96 & 37.7 &  36.8   &&  8.74    & 8.26 & 24.6 & 23.3 \\
    \texttt{4} & PREVALENT~\cite{hao2020towards}
        && 10.24     & 6.32 & 45.8 &  44.2   &&  11.91   & 7.28 & 33.8 & 31.1 \\
    \texttt{5} & RelGraph~\cite{hong2020language}
        && 9.19      & 7.10 & 36.6 &  35.3   &&  9.26     & 7.79 & 29.0 & 27.4 \\
    \hline
    &\textit{Exploration-based Agents} \\
    \texttt{6} & FAST-Short~\cite{ke2019tactical}
        && 14.70     & 5.52 & 52.3 &  46.3   && 22.89    & 6.78 & 36.8 & 26.4 \\
    \texttt{7} & Active VLN~\cite{wang2020active}
        && 24.35     & 6.48    & 43.1    & 29.3    && 19.40 & 7.08 & 32.2 & 21.2 \\
    \texttt{8} & SSM~\cite{wang2021structured}
        && 20.13     & 6.14 & 49.4 &  40.9   && 26.64    & 6.70 & 39.8 & 26.1 \\
    \hline
    \texttt{9} & VLN$\protect\CircleArrowright$BERT~\cite{hong2021vln}
        && 12.29     & 5.80 & 47.9 & 44.2    &&  13.00   & 6.47 & 39.3 & 35.0 \\
    \rowcolor{gray!10} \texttt{10} & Ours (VLN$\protect\CircleArrowright$BERT-based, w/o E2E)
        && 11.92     & 5.60 & 49.1 &  \textbf{45.7} && 12.95    & 6.19 & 42.3 & \textbf{37.7} \\
    \rowcolor{gray!10} \texttt{11} & Ours (VLN$\protect\CircleArrowright$BERT-based, w/ E2E)
        && 19.28     & \textbf{5.56} & \textbf{50.7} &  38.1 && 23.02    & \textbf{6.13} & \textbf{44.7} & 29.7 \\
    \texttt{12} & HAMT~\cite{chen2021history}
        && 11.79     & 5.65 & 49.1 & 46.2   &&  12.98   & 6.33 & 41.7 & 37.6 \\
    \rowcolor{gray!10} \texttt{13} & Ours (HAMT-based, w/o E2E)
        && 12.44     & 5.36 & 52.3 &  \textbf{49.3} && 12.91    & 6.10 & 43.5 & \textbf{39.5}  \\
    \rowcolor{gray!10} \texttt{14} & Ours (HAMT-based, w/ E2E)
        && 27.08     & \textbf{5.21} & \textbf{54.2} &  34.7 && 28.31    & \textbf{6.05} & \textbf{44.6} & 25.9 \\
    \hline \hline
  \end{tabular}}
\end{center}
\caption{Comparison of VLN agents on R2R-ULN validation set, including all three levels $L_1, L_2, L_3$. Rows 1-5 are results from greedy-decoding agents without back-tracking. Rows 6-8 are results from agents with back-tracking or graph-based search. Note that the metric values shown here are averaged across three levels of underspecification.}
\label{tab:uln_result}
\end{table*}

\paragraph{Dataset} R2R \cite{anderson2018vision} contains over 14k instructions for training, 1k for validation seen environments (Val-seen), 2.3k for validation unseen (Val-unseen). As for R2R-ULN, we select around 1600 longest instructions from the R2R validation set as $L_0$. We assign three different annotators to simplify each $L_0$ and filter low-quality samples. R2R-ULN includes 3132 instructions for Val-seen, 6714 for Val-unseen, and 4198 test unseen. We train the agent and other modules on R2R (train split) only and evaluate our system on R2R and R2R-ULN without re-training. To maintain the ratio between the training and validation set, we use a subset of 2k Val-seen and 4.5k Val-unseen from the full R2R-ULN for evaluation as default if not specified. We randomly sample 30\% L1, 70\% L2 and 100\% L3 from the full set based on the preference results (Table~\ref{tab:human_verify}). We also report the evaluation results in the full set in Appendix.

\paragraph{Evaluation Metrics} We evaluate the navigation performance using the standard metrics of R2R: Trajectory Length~(TL): the agent's navigation path in meters; Navigation Error~(NE $\downarrow$): the average distance between the goal and agent's final location; Success Rate~(SR $\uparrow$): the ratio of trials that end within 3 meters to the overall target trials; Success weighted by inverse Path Length~(SPL $\uparrow$)~\cite{anderson2018evaluation}.

\paragraph{Implementation Details}\label{para:implement}
We adopt full instructions as low-level samples (R2R) and the last sentences of instructions as high-level samples (R2R-last). We train the classification module and the agents with these two training sets. For uncertainty estimation training, we applied the chunking function \cite{hong2020sub} to randomly drop sub-instructions from R2R and create pseudo-underspecified instructions as inputs to a trained agent. At each step, if the agent's action is different from the teacher's action, the uncertainty ground truth label is 1, else 0. During inference, an uncertainty score over 0.5 will initiate multi-step lookahead. We also limit the lookahead to at most three times for performance benefit. We explain this choice in Sec. \ref{subsec:ab_exp}. 

We train the classification and the agent with a low-level text encoder for 300,000 iterations with a learning rate of 1e-2 and 1e-5. The agent is trained on a mix of imitation learning, and A2C \cite{mnih2016asynchronous}, the same as the baselines. Then we train the high-level text encoder with other parameters fixed for 10,000 iterations. Finally, for the E2E module, we train the uncertainty estimation network with a learning rate 1e-4 for 10 epochs. We directly adopt the trained agent as the explorer for efficiency. All training stages are done on a single GPU with AdamW optimizer \cite{loshchilov2018decoupled}. For E2E, we set $\gamma$ as 1.2 and $K$ as 1 for the best performance and balance between SR and SPL.

\begin{table}[t]
  \begin{center}
  \resizebox{1.0\linewidth}{!}{
  \begin{tabular}{l|c|c|c|c|c|c}
    \hline \hline
    \multicolumn{1}{c}{\multirow{3}{*}{Methods}} &\multicolumn{3}{|c|}{R2R-ULN Val-Seen} & \multicolumn{3}{c}{R2R-ULN Val-Unseen} \\
    \cline{2-7} &
    \multicolumn{1}{c|}{L1} & \multicolumn{1}{c|}{L2} & \multicolumn{1}{c|}{L3} & 
    \multicolumn{1}{c|}{L1} & \multicolumn{1}{c|}{L2} & \multicolumn{1}{c}{L3} \\
    \cline{2-7} &
    \multicolumn{1}{c|}{SR$\uparrow$} & 
    \multicolumn{1}{c|}{SR$\uparrow$} &
    \multicolumn{1}{c|}{SR$\uparrow$} &
    \multicolumn{1}{c|}{SR$\uparrow$} &
    \multicolumn{1}{c|}{SR$\uparrow$} &
    \multicolumn{1}{c}{SR$\uparrow$} \\
    \hline \hline
    VLN$\protect\CircleArrowright$BERT
        & 62.4 & 55.6 & 38.5 & 50.2 & 44.2 & 32.7 \\
    Ours (w/o E2E)
        & 62.4 & 54.8 & 41.3 & 50.5 & 45.4 & 37.8 \\
    Ours (w/ E2E)
        & \textbf{62.7} & \textbf{56.6} & \textbf{43.2} & \textbf{55.7} & \textbf{47.6} & \textbf{39.5} \\  
    \hline
    HAMT
        & \textbf{64.7} & \textbf{57.6} & 38.7 & \textbf{56.4} & 44.6 & 35.1 \\
    Ours (w/o E2E)
        & 64.4 & 55.9 & \textbf{46.3} & 56.0 & 44.9 & 38.8 \\
    Ours (w/ E2E)
        & 62.4 & 56.8 & 43.6 & 55.5 & \textbf{47.7} & \textbf{39.5} \\  
    \hline \hline
  \end{tabular}}
\end{center}
\caption{Performance breakdown by different levels of underspecification on R2R-ULN.}
\label{tab:uln_breakdown}
\end{table}

\subsection{Main Results}
\label{subsec:main_results}
\paragraph{Comparison with SOTA Agents} We first compare our GSS method with the state-of-the-art (SOTA) model HAMT \cite{chen2021history}, and a strong baseline VLN$\protect\CircleArrowright$BERT. We mainly consider three training sets for the baselines: R2R only, R2R-last only, and R2R+R2R-last. Table~\ref{tab:dual_encoder_result} shows that training on a single granularity inevitably degrades the agent's performance on the other level. A mixture of these two levels shows a compromised performance across R2R and R2R-ULN validation sets. In contrast, our method can achieve 3.8\% absolute SR improvement for R2R-ULN $L_3$ while maintaining the performance on R2R. Our GSS only requires 30\% additional training iterations, 25\% - 38\% extra parameters, and can achieve slightly better performance.

\paragraph{Greedy-Decoding Agents} As is shown in the top section (Row 1-5) of Table~\ref{tab:uln_result}, greedy-decoding agents generally struggle with R2R-ULN instructions. These agents' SR and SPL values generally drop relatively by 40-50\%, indicating a potential performance degradation and risk due to language variations when we deploy these models in more realistic household environments. Note that VLN$\protect\CircleArrowright$BERT and HAMT are more robust with only a relatively 30\% decrease in SR. This may attribute to better transformer architectures and initialization from large-scale pre-trained models.

\paragraph{Exploration-based Agents} We select three exploration-based agents: FAST-Short \cite{ke2019tactical}, Active VLN \cite{wang2020active}, SSM \cite{wang2021structured}. Table \ref{tab:uln_result} shows that FAST and SSM navigate longer trajectories and achieve better SR. They are even better than VLN$\protect\CircleArrowright$BERT and HAMT on Val-seen though they underperform these SOTA agents on R2R. 

Our system outperforms the corresponding baselines by around 2\% in SR and 1.6\% in SPL without exploration. With the E2E module, our system gains additional improvement by sacrificing the trajectory length, resulting in lower SPL values. The improvement from GSS and E2E is consistent, as shown in Table \ref{tab:uln_breakdown}. Since the target level of GSS is $L_3$, only $L_3$ SR is improved without E2E. On the other hand, E2E can improve performance across all levels by estimating grounding uncertainty and looking ahead of frontiers.

\paragraph{Comparison on R2R} Table~\ref{tab:dual_encoder_result} shows the evaluation results on R2R Val-unseen. Our classification module and GSS can achieve similar or better performance than the baselines. We disable E2E since ULN does not encourage exploration for $L_0$, but we also show the performance with E2E in Appendix \ref{sec:supp_exp} Table \ref{tab:r2r_result_appendix}. E2E improves the SR by 1.4\% by sacrificing path length for VLN$\protect\CircleArrowright$BERT while decreasing SR by 1.8\% for HAMT. This is potentially due to inaccurate uncertainty estimation and excessive exploration for full instructions. Considering the overall performance in R2R and R2R-ULN, our method still improves the SR by an  absolute 1-3 \% (see Appendix \ref{app:ablation}).

\subsection{Ablation Studies}

\begin{table}[t]
  \begin{center}
  \resizebox{\linewidth}{!}{
  \begin{tabular}{cccc c c c c}
    \hline \hline
    \multicolumn{4}{c}{Components} && \multicolumn{3}{c}{R2R-ULN Val-Unseen}\\
    \cline{1-4} \cline{6-8}
    \multicolumn{1}{c}{\multirow{2}{*}{Classify Instr.}} & \multicolumn{1}{c}{\multirow{2}{*}{GSS}} & \multicolumn{2}{c}{E2E} && \multicolumn{1}{c}{L1} & \multicolumn{1}{c}{L2} & \multicolumn{1}{c}{L3}\\
    \cline{3-4} \cline{6-8} 
     &  & \multicolumn{1}{c}{Lookahead} & \multicolumn{1}{c}{State Freeze} &&
    \multicolumn{1}{c}{SR$\uparrow$} & 
    \multicolumn{1}{c}{SR$\uparrow$} &  
    \multicolumn{1}{c}{SR$\uparrow$} \\
    \hline
     & & & && 50.8  & 44.1 & 32.7 \\
     & \checkmark & & && 50.8 & 44.1 & 37.8 \\
     \checkmark & \checkmark & & && 51.0 & 45.1 & 37.8\\
     & & \checkmark & && 51.8 & 46.7 & 34.5 \\
     & & \checkmark & \checkmark && \textbf{52.6} & 46.6 & 35.9 \\
     \checkmark & \checkmark & \checkmark & & & 52.0 & 46.8 & 37.8 \\
     \checkmark & \checkmark & \checkmark & \checkmark && \textbf{52.7} & \textbf{47.4} & \textbf{39.5}\\
    \hline \hline
  \end{tabular}
  }
\end{center}
\caption{Ablation study for our framework components. We use VLN$\protect\CircleArrowright$BERT as the baseline and run experiments on full R2R-ULN Val-unseen. We use the ground truth class for GSS without a classification module.}
\label{tab:ablation}
\end{table}

\subsubsection{Component Analysis}
We demonstrate the effectiveness of the GSS and E2E module in our framework in Table~\ref{tab:ablation}.
Adding GSS brings the most significant SR gain in $L_3$ by 5.1\% percent. This gain is intuitive as the high-level sub-network is trained on coarse-grained instructions only. Adding the classification module harms the performance for $L_1$ but improves it for $L_2$. That is because of the 85\% accuracy and potentially some $L_2$ instructions being too short. The high-level sub-network is thus more suitable for these $L_2$ instructions. Secondly, adding the lookahead heuristic improves the SR by 1-2\% consistently across all levels. The state freeze trick is also beneficial as it further improves the SR by 1\% for $L_1, L_3$. Finally, we verify that gains from GSS and E2E are supplementary to each other. The last two rows show that adding lookahead on top of GSS can improve 1-2\% SR on $L_1, L_2$ but cannot help with $L_3$. With state freeze, we observe an additional 2\% SR gain in $L_3$. The potential reason is that the lookahead heuristic has overlapping benefits with GSS, but state freeze is a complementary trick for high-level sub-network. 

\begin{table}[t]
  \begin{center}
  \resizebox{1.0\linewidth}{!}{
  \begin{tabular}{c|c|cc|cc|cc}
    \hline \hline
    \multicolumn{1}{c|}{\multirow{3}{*}{Base}} & \multicolumn{1}{c|}{\multirow{3}{*}{\makecell[c]{Uncertainty\\Threshold}}} & \multicolumn{6}{c}{R2R-ULN Val-Unseen} \\
    \cline{3-8} &
    & \multicolumn{2}{c|}{L1} & \multicolumn{2}{c|}{L2} & \multicolumn{2}{c}{L3} \\
    \cline{3-8} &
    & \multicolumn{1}{c}{SR$\uparrow$} & \multicolumn{1}{c|}{SPL$\uparrow$} &
    \multicolumn{1}{c}{SR$\uparrow$} & \multicolumn{1}{c|}{SPL$\uparrow$} &
    \multicolumn{1}{c}{SR$\uparrow$} & \multicolumn{1}{c}{SPL$\uparrow$} \\
    \hline \hline
    \multicolumn{1}{c|}{\multirow{5}{*}{VLN\CircleArrowright BERT}}
    & 0.00
        & 53.2 & 13.7 & 46.5 & 12.1 & 38.0 & 9.2 \\
    & 0.25
        & 53.9 & 25.7 & 48.5 & 23.8 & 39.9 & 18.2 \\
    & \textcolor{red}{0.50}
        & 52.6 & 36.9 & 47.4 & 32.5 & 39.5 & 25.6 \\
    & 0.75 
        & 51.5 & 44.0 & 45.7 & 38.6 & 38.5 & 31.3 \\  
    & 1.00
        & 51.0 & 45.6 & 45.1 & 40.4 & 37.8 & 33.5 \\
    \hline \hline
  \end{tabular}}
\end{center}
\caption{Ablation study on uncertainty threshold. We mark the value with the best SR-SPL trade-off in red.}
\label{tab:ablation2}
\end{table}

\subsubsection{Exploration Threshold}
As is mentioned in Section \ref{subsec:e2e}, the E2E module initiates an exploration when the uncertainty score $s^{\text{uncert}}$ exceeds a threshold. We investigate the effect of this threshold value on the evaluation results in Table \ref{tab:ablation2}. A threshold value of 1.0 indicates no exploration at all. Decreasing the threshold imposes more explorations and improves SR by 2-3\% across all levels at the expense of lower SPL. When the value goes below 0.5, the improvement in SR is marginal, while the reduction in SPL is significant. Therefore, we select 0.5 as our threshold for the best SR-SPL trade-off in practice.

\subsubsection{Exploration Accuracy} \label{subsec:ab_exp}

\begin{figure}[t]
    \centering
    \includegraphics[width=\linewidth]{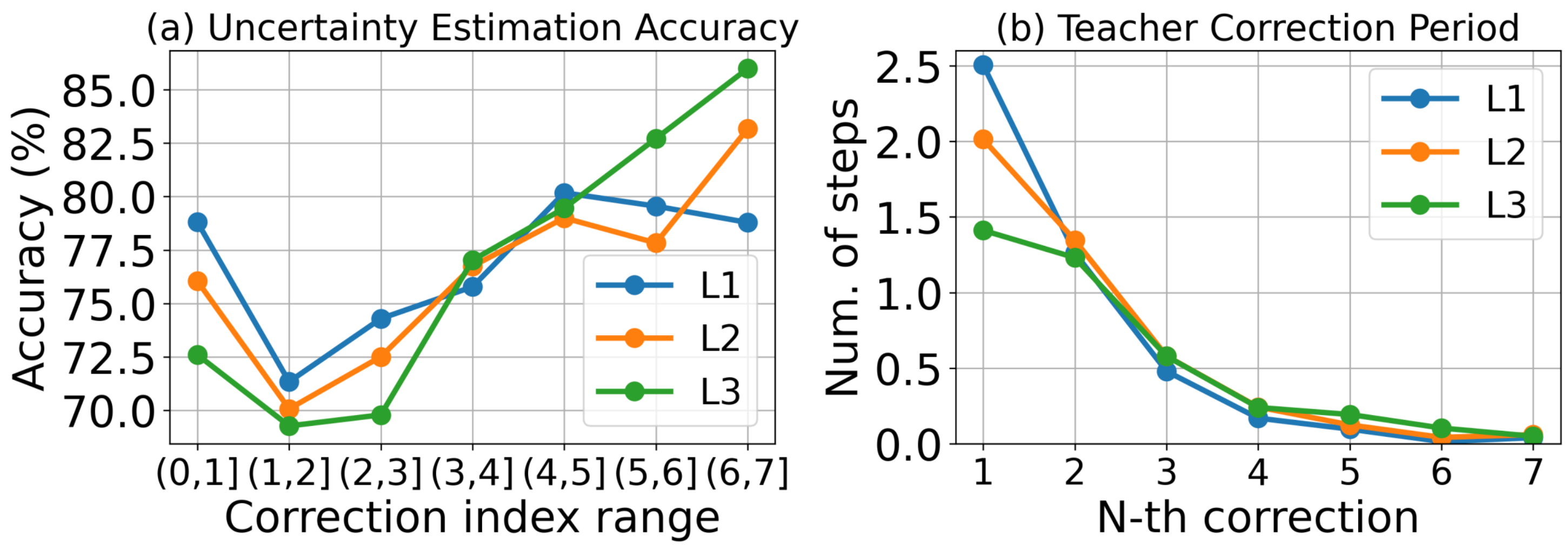}
    \caption{(a) Average estimation accuracy for steps between two consecutive corrections. (b) The average number of steps before taking N-th correction. }
    \label{fig:uncert}
\end{figure}

\paragraph{Uncertainty Estimation} Beyond success-based metrics, the accuracy of uncertainty estimation is an important indicator of whether the exploration steps are necessary. Given an underspecified instruction, we measure the accuracy with teacher corrections, i.e., correcting the agent's action when it deviates from the teacher's action. Figure~\ref{fig:uncert}(a) shows the accuracy for steps between two consecutive corrections. Interestingly, the accuracy drops to a minimum between the first and second correction and increases as the agent moves forward. Despite that the estimation becomes most inaccurate when the agent makes its second or third false decision, we will show that these steps are critical ones to be adjusted. 

\paragraph{Early Exploration} One interesting question that arises from the above observation is: \textit{should we seize explorations for the first several uncertainties and encourage explorations in later steps?} The answer can be revealed by investigating the number of steps between two teacher corrections. Surprisingly, Figure~\ref{fig:uncert}(b) indicates that $Ti$ deviates from $T_0$ at almost every step after the third correction. Assuming that our one-step lookahead is perfect as the teacher correction, the phenomenon implies that our agent has to make an exploration for every single step after the third exploration. However, such behavior is undesirable as it excessively harms navigation efficiency. Fig. \ref{fig:uncert} together implies that agents must rely on early explorations instead of later ones. Otherwise, the system error accumulates exponentially and becomes intractable eventually. Therefore, our agent relies on early explorations even though the uncertainty estimation accuracy is relatively defective.

\section{Conclusion}
In this work, we consider a new setting, Underspecified vision-and-Language Navigation~(ULN). We collected a large-scale evaluation set, R2R-ULN, with multi-level underspecified instructions. We show that ULN is a reasonable and practical setting. ULN presents a novel direction where explorations are necessary and justifiable. As a first step towards ULN, we propose two novel components to build a VLN framework, Granularity Specific Sub-network (GSS) and Exploitation-to-Exploration (E2E). Experimental results show that GSS and E2E effectively mitigate the performance gap across all levels of instruction. Finally, we believe that ULN is a more challenging setting for future VLN development, and our framework can be further improved with more sophisticated policy design or language grounding models.

\section{Limitations}
In this study, we only cover Vision-Language Navigation datasets with English instructions. Instructions in other languages may characterize different types of ambiguity or underspecification. Thus, expanding the datasets to multi-lingual ULN based on datasets like RxR is essential. Secondly, we only consider indoor environments where the instructions are generally shorter than outdoor ones due to shorter path lengths. However, the phenomenon of underspecification can also be expected outdoors, accompanied by other modalities such as hand gestures or hand sketches. We simply assumed that underspecification is a more ubiquitous phenomenon in the indoor than the outdoor environment, which may be overturned from additional surveys or experiments. In the future, we hope to expand our work to multi-lingual instructions and outdoor environments and combine it with more modalities. 

\section{Ethical Considerations}
For data collection and verification on Amazon Mechanical Turk, we select annotators from English-speaking countries, including the US, CA, UK, AU, and NZ. Each HIT for instruction simplification takes around 1.5 minutes on average to accomplish, and we pay each submitted HIT with 0.4 US dollars, resulting in an hourly payment of 16 US dollars. As for the instruction following, each HIT takes around 2 minutes to accomplish, and we pay each HIT 0.5 US dollars per HIT, resulting in an hourly payment of 15 US dollars. In addition, we award each successful navigation attempt with 0.3 US dollars for high-quality verification. As for the preference assessment, each HIT takes around 1 minute to accomplish, and we pay each HIT 0.3 US dollars, resulting in an hourly payment of 18 US dollars. 

\section{Acknowledgment}
We would like to thank the Robert N. Noyce Trust for their generous gift to the University of California via the Noyce Initiative. The work is also partially funded by an unrestricted gift from Google. The writers' opinions and conclusions in this publication are their own and should not be construed as representing the sponsors' official policy, expressed or inferred.

\bibliography{anthology,custom}

\begin{thebibliography}{57}
\expandafter\ifx\csname natexlab\endcsname\relax\def\natexlab#1{#1}\fi

\bibitem[{Agrawal et~al.(2018)Agrawal, Batra, Parikh, and
  Kembhavi}]{agrawal2018don}
Aishwarya Agrawal, Dhruv Batra, Devi Parikh, and Aniruddha Kembhavi. 2018.
\newblock Don't just assume; look and answer: Overcoming priors for visual
  question answering.
\newblock In \emph{Proceedings of the IEEE conference on computer vision and
  pattern recognition}, pages 4971--4980.

\bibitem[{Akula et~al.(2020)Akula, Gella, Al-Onaizan, Zhu, and
  Reddy}]{akula2020words}
Arjun Akula, Spandana Gella, Yaser Al-Onaizan, Song-chun Zhu, and Siva Reddy.
  2020.
\newblock Words aren’t enough, their order matters: On the robustness of
  grounding visual referring expressions.
\newblock In \emph{Proceedings of the 58th Annual Meeting of the Association
  for Computational Linguistics}, pages 6555--6565.

\bibitem[{Alayrac et~al.(2020)Alayrac, Recasens, Schneider, Arandjelovi{\'c},
  Ramapuram, De~Fauw, Smaira, Dieleman, and Zisserman}]{alayrac2020self}
Jean-Baptiste Alayrac, Adria Recasens, Rosalia Schneider, Relja
  Arandjelovi{\'c}, Jason Ramapuram, Jeffrey De~Fauw, Lucas Smaira, Sander
  Dieleman, and Andrew Zisserman. 2020.
\newblock Self-supervised multimodal versatile networks.
\newblock \emph{Advances in Neural Information Processing Systems}, 33:25--37.

\bibitem[{Anderson et~al.(2018{\natexlab{a}})Anderson, Chang, Chaplot,
  Dosovitskiy, Gupta, Koltun, Kosecka, Malik, Mottaghi, Savva
  et~al.}]{anderson2018evaluation}
Peter Anderson, Angel Chang, Devendra~Singh Chaplot, Alexey Dosovitskiy,
  Saurabh Gupta, Vladlen Koltun, Jana Kosecka, Jitendra Malik, Roozbeh
  Mottaghi, Manolis Savva, et~al. 2018{\natexlab{a}}.
\newblock On evaluation of embodied navigation agents.
\newblock \emph{arXiv preprint arXiv:1807.06757}.

\bibitem[{Anderson et~al.(2018{\natexlab{b}})Anderson, Wu, Teney, Bruce,
  Johnson, S{\"u}nderhauf, Reid, Gould, and Van
  Den~Hengel}]{anderson2018vision}
Peter Anderson, Qi~Wu, Damien Teney, Jake Bruce, Mark Johnson, Niko
  S{\"u}nderhauf, Ian Reid, Stephen Gould, and Anton Van Den~Hengel.
  2018{\natexlab{b}}.
\newblock Vision-and-language navigation: Interpreting visually-grounded
  navigation instructions in real environments.
\newblock In \emph{Proceedings of the IEEE Conference on Computer Vision and
  Pattern Recognition}, pages 3674--3683.

\bibitem[{Antol et~al.(2015)Antol, Agrawal, Lu, Mitchell, Batra, Zitnick, and
  Parikh}]{antol2015vqa}
Stanislaw Antol, Aishwarya Agrawal, Jiasen Lu, Margaret Mitchell, Dhruv Batra,
  C~Lawrence Zitnick, and Devi Parikh. 2015.
\newblock Vqa: Visual question answering.
\newblock In \emph{Proceedings of the IEEE international conference on computer
  vision}, pages 2425--2433.

\bibitem[{Bisk et~al.(2016)Bisk, Yuret, and Marcu}]{bisk2016natural}
Yonatan Bisk, Deniz Yuret, and Daniel Marcu. 2016.
\newblock Natural language communication with robots.
\newblock In \emph{Proceedings of the 2016 Conference of the North American
  Chapter of the Association for Computational Linguistics: Human Language
  Technologies}, pages 751--761.

\bibitem[{Chang et~al.(2017)Chang, Dai, Funkhouser, Halber, Niebner, Savva,
  Song, Zeng, and Zhang}]{chang2017matterport3d}
Angel Chang, Angela Dai, Thomas Funkhouser, Maciej Halber, Matthias Niebner,
  Manolis Savva, Shuran Song, Andy Zeng, and Yinda Zhang. 2017.
\newblock Matterport3d: Learning from rgb-d data in indoor environments.
\newblock In \emph{2017 International Conference on 3D Vision (3DV)}, pages
  667--676. IEEE.

\bibitem[{Chen et~al.(2019)Chen, Suhr, Misra, Snavely, and
  Artzi}]{chen2019touchdown}
Howard Chen, Alane Suhr, Dipendra Misra, Noah Snavely, and Yoav Artzi. 2019.
\newblock Touchdown: Natural language navigation and spatial reasoning in
  visual street environments.
\newblock In \emph{Proceedings of the IEEE/CVF Conference on Computer Vision
  and Pattern Recognition}, pages 12538--12547.

\bibitem[{Chen et~al.(2021{\natexlab{a}})Chen, Guhur, Schmid, and
  Laptev}]{chen2021history}
Shizhe Chen, Pierre-Louis Guhur, Cordelia Schmid, and Ivan Laptev.
  2021{\natexlab{a}}.
\newblock History aware multimodal transformer for vision-and-language
  navigation.
\newblock \emph{Advances in Neural Information Processing Systems}, 34.

\bibitem[{Chen et~al.(2022)Chen, Guhur, Tapaswi, Schmid, and
  Laptev}]{chen2022think}
Shizhe Chen, Pierre-Louis Guhur, Makarand Tapaswi, Cordelia Schmid, and Ivan
  Laptev. 2022.
\newblock Think global, act local: Dual-scale graph transformer for
  vision-and-language navigation.
\newblock In \emph{Proceedings of the IEEE/CVF Conference on Computer Vision
  and Pattern Recognition}, pages 16537--16547.

\bibitem[{Chen et~al.(2021{\natexlab{b}})Chen, Li, Kong, Kei, Zhu, Gao, Zhu,
  and Huang}]{chen2021yourefit}
Yixin Chen, Qing Li, Deqian Kong, Yik~Lun Kei, Song-Chun Zhu, Tao Gao, Yixin
  Zhu, and Siyuan Huang. 2021{\natexlab{b}}.
\newblock Yourefit: Embodied reference understanding with language and gesture.
\newblock In \emph{Proceedings of the IEEE/CVF International Conference on
  Computer Vision}, pages 1385--1395.

\bibitem[{Cirik et~al.(2018)Cirik, Morency, and
  Berg-Kirkpatrick}]{cirik2018visual}
Volkan Cirik, Louis-Philippe Morency, and Taylor Berg-Kirkpatrick. 2018.
\newblock Visual referring expression recognition: What do systems actually
  learn?
\newblock In \emph{Proceedings of the 2018 Conference of the North American
  Chapter of the Association for Computational Linguistics: Human Language
  Technologies, Volume 2 (Short Papers)}, pages 781--787.

\bibitem[{Deng et~al.(2020)Deng, Narasimhan, and
  Russakovsky}]{deng2020evolving}
Zhiwei Deng, Karthik Narasimhan, and Olga Russakovsky. 2020.
\newblock Evolving graphical planner: Contextual global planning for
  vision-and-language navigation.
\newblock \emph{Advances in Neural Information Processing Systems},
  33:20660--20672.

\bibitem[{Dosovitskiy et~al.(2020)Dosovitskiy, Beyer, Kolesnikov, Weissenborn,
  Zhai, Unterthiner, Dehghani, Minderer, Heigold, Gelly
  et~al.}]{dosovitskiy2020image}
Alexey Dosovitskiy, Lucas Beyer, Alexander Kolesnikov, Dirk Weissenborn,
  Xiaohua Zhai, Thomas Unterthiner, Mostafa Dehghani, Matthias Minderer, Georg
  Heigold, Sylvain Gelly, et~al. 2020.
\newblock An image is worth 16x16 words: Transformers for image recognition at
  scale.
\newblock In \emph{International Conference on Learning Representations}.

\bibitem[{Fang et~al.(2022)Fang, Ye, Li, Feng, and Wang}]{fang2022stemm}
Qingkai Fang, Rong Ye, Lei Li, Yang Feng, and Mingxuan Wang. 2022.
\newblock Stemm: Self-learning with speech-text manifold mixup for speech
  translation.
\newblock In \emph{Proceedings of the 60th Annual Meeting of the Association
  for Computational Linguistics (Volume 1: Long Papers)}, pages 7050--7062.

\bibitem[{Fried et~al.(2018)Fried, Hu, Cirik, Rohrbach, Andreas, Morency,
  Berg-Kirkpatrick, Saenko, Klein, and Darrell}]{fried2018speaker}
Daniel Fried, Ronghang Hu, Volkan Cirik, Anna Rohrbach, Jacob Andreas,
  Louis-Philippe Morency, Taylor Berg-Kirkpatrick, Kate Saenko, Dan Klein, and
  Trevor Darrell. 2018.
\newblock Speaker-follower models for vision-and-language navigation.
\newblock In \emph{Proceedings of the 32nd International Conference on Neural
  Information Processing Systems}, pages 3318--3329.

\bibitem[{Fu et~al.(2020)Fu, Wang, Peterson, Grafton, Eckstein, and
  Wang}]{fu2020counterfactual}
Tsu-Jui Fu, Xin~Eric Wang, Matthew~F Peterson, Scott~T Grafton, Miguel~P
  Eckstein, and William~Yang Wang. 2020.
\newblock Counterfactual vision-and-language navigation via adversarial path
  sampler.
\newblock In \emph{European Conference on Computer Vision}, pages 71--86.
  Springer.

\bibitem[{Gu et~al.(2022)Gu, Stefani, Wu, Thomason, and Wang}]{gu2022vision}
Jing Gu, Eliana Stefani, Qi~Wu, Jesse Thomason, and Xin Wang. 2022.
\newblock Vision-and-language navigation: A survey of tasks, methods, and
  future directions.
\newblock In \emph{Proceedings of the 60th Annual Meeting of the Association
  for Computational Linguistics (Volume 1: Long Papers)}, pages 7606--7623.

\bibitem[{Guhur et~al.(2021)Guhur, Tapaswi, Chen, Laptev, and
  Schmid}]{guhur2021airbert}
Pierre-Louis Guhur, Makarand Tapaswi, Shizhe Chen, Ivan Laptev, and Cordelia
  Schmid. 2021.
\newblock Airbert: In-domain pretraining for vision-and-language navigation.
\newblock In \emph{Proceedings of the IEEE/CVF International Conference on
  Computer Vision}, pages 1634--1643.

\bibitem[{Hao et~al.(2020)Hao, Li, Li, Carin, and Gao}]{hao2020towards}
Weituo Hao, Chunyuan Li, Xiujun Li, Lawrence Carin, and Jianfeng Gao. 2020.
\newblock Towards learning a generic agent for vision-and-language navigation
  via pre-training.
\newblock In \emph{Proceedings of the IEEE/CVF Conference on Computer Vision
  and Pattern Recognition}, pages 13137--13146.

\bibitem[{He et~al.(2021)He, Huang, Wu, Yang, An, Sima, and
  Wang}]{he2021landmark}
Keji He, Yan Huang, Qi~Wu, Jianhua Yang, Dong An, Shuanglin Sima, and Liang
  Wang. 2021.
\newblock Landmark-rxr: Solving vision-and-language navigation with
  fine-grained alignment supervision.
\newblock \emph{Advances in Neural Information Processing Systems},
  34:652--663.

\bibitem[{Hong et~al.(2020{\natexlab{a}})Hong, Rodriguez, Wu, and
  Gould}]{hong2020sub}
Yicong Hong, Cristian Rodriguez, Qi~Wu, and Stephen Gould. 2020{\natexlab{a}}.
\newblock Sub-instruction aware vision-and-language navigation.
\newblock In \emph{Proceedings of the 2020 Conference on Empirical Methods in
  Natural Language Processing (EMNLP)}, pages 3360--3376.

\bibitem[{Hong et~al.(2020{\natexlab{b}})Hong, Rodriguez-Opazo, Qi, Wu, and
  Gould}]{hong2020language}
Yicong Hong, Cristian Rodriguez-Opazo, Yuankai Qi, Qi~Wu, and Stephen Gould.
  2020{\natexlab{b}}.
\newblock Language and visual entity relationship graph for agent navigation.
\newblock \emph{arXiv preprint arXiv:2010.09304}.

\bibitem[{Hong et~al.(2021)Hong, Wu, Qi, Rodriguez-Opazo, and
  Gould}]{hong2021vln}
Yicong Hong, Qi~Wu, Yuankai Qi, Cristian Rodriguez-Opazo, and Stephen Gould.
  2021.
\newblock Vln bert: A recurrent vision-and-language bert for navigation.
\newblock In \emph{Proceedings of the IEEE/CVF Conference on Computer Vision
  and Pattern Recognition}, pages 1643--1653.

\bibitem[{Jain et~al.(2019)Jain, Magalhaes, Ku, Vaswani, Ie, and
  Baldridge}]{jain2019stay}
Vihan Jain, Gabriel Magalhaes, Alexander Ku, Ashish Vaswani, Eugene Ie, and
  Jason Baldridge. 2019.
\newblock Stay on the path: Instruction fidelity in vision-and-language
  navigation.
\newblock \emph{arXiv preprint arXiv:1905.12255}.

\bibitem[{Johnson et~al.(2017)Johnson, Hariharan, Van Der~Maaten, Fei-Fei,
  Lawrence~Zitnick, and Girshick}]{johnson2017clevr}
Justin Johnson, Bharath Hariharan, Laurens Van Der~Maaten, Li~Fei-Fei,
  C~Lawrence~Zitnick, and Ross Girshick. 2017.
\newblock Clevr: A diagnostic dataset for compositional language and elementary
  visual reasoning.
\newblock In \emph{Proceedings of the IEEE conference on computer vision and
  pattern recognition}, pages 2901--2910.

\bibitem[{Kazemzadeh et~al.(2014)Kazemzadeh, Ordonez, Matten, and
  Berg}]{kazemzadeh2014referitgame}
Sahar Kazemzadeh, Vicente Ordonez, Mark Matten, and Tamara Berg. 2014.
\newblock Referitgame: Referring to objects in photographs of natural scenes.
\newblock In \emph{Proceedings of the 2014 conference on empirical methods in
  natural language processing (EMNLP)}, pages 787--798.

\bibitem[{Ke et~al.(2019)Ke, Li, Bisk, Holtzman, Gan, Liu, Gao, Choi, and
  Srinivasa}]{ke2019tactical}
Liyiming Ke, Xiujun Li, Yonatan Bisk, Ari Holtzman, Zhe Gan, Jingjing Liu,
  Jianfeng Gao, Yejin Choi, and Siddhartha Srinivasa. 2019.
\newblock Tactical rewind: Self-correction via backtracking in
  vision-and-language navigation.
\newblock In \emph{Proceedings of the IEEE/CVF Conference on Computer Vision
  and Pattern Recognition}, pages 6741--6749.

\bibitem[{Koh et~al.(2021)Koh, Lee, Yang, Baldridge, and
  Anderson}]{koh2021pathdreamer}
Jing~Yu Koh, Honglak Lee, Yinfei Yang, Jason Baldridge, and Peter Anderson.
  2021.
\newblock Pathdreamer: A world model for indoor navigation.
\newblock In \emph{Proceedings of the IEEE/CVF International Conference on
  Computer Vision}, pages 14738--14748.

\bibitem[{Kolve et~al.(2017)Kolve, Mottaghi, Han, VanderBilt, Weihs, Herrasti,
  Gordon, Zhu, Gupta, and Farhadi}]{kolve2017ai2}
Eric Kolve, Roozbeh Mottaghi, Winson Han, Eli VanderBilt, Luca Weihs, Alvaro
  Herrasti, Daniel Gordon, Yuke Zhu, Abhinav Gupta, and Ali Farhadi. 2017.
\newblock Ai2-thor: An interactive 3d environment for visual ai.
\newblock \emph{arXiv preprint arXiv:1712.05474}.

\bibitem[{Ku et~al.(2020)Ku, Anderson, Patel, Ie, and Baldridge}]{ku2020room}
Alexander Ku, Peter Anderson, Roma Patel, Eugene Ie, and Jason Baldridge. 2020.
\newblock Room-across-room: Multilingual vision-and-language navigation with
  dense spatiotemporal grounding.
\newblock In \emph{Proceedings of the 2020 Conference on Empirical Methods in
  Natural Language Processing (EMNLP)}, pages 4392--4412.

\bibitem[{Li et~al.(2021)Li, Tan, and Bansal}]{li2021improving}
Jialu Li, Hao Tan, and Mohit Bansal. 2021.
\newblock Improving cross-modal alignment in vision language navigation via
  syntactic information.
\newblock In \emph{Proceedings of the 2021 Conference of the North American
  Chapter of the Association for Computational Linguistics: Human Language
  Technologies}, pages 1041--1050.

\bibitem[{Li et~al.(2022)Li, Tan, and Bansal}]{li2022envedit}
Jialu Li, Hao Tan, and Mohit Bansal. 2022.
\newblock Envedit: Environment editing for vision-and-language navigation.
\newblock In \emph{Proceedings of the IEEE/CVF Conference on Computer Vision
  and Pattern Recognition}, pages 15407--15417.

\bibitem[{Lin et~al.(2021)Lin, Zhu, Long, Liang, Ye, and
  Lin}]{lin2021adversarial}
Bingqian Lin, Yi~Zhu, Yanxin Long, Xiaodan Liang, Qixiang Ye, and Liang Lin.
  2021.
\newblock Adversarial reinforced instruction attacker for robust
  vision-language navigation.
\newblock \emph{arXiv preprint arXiv:2107.11252}.

\bibitem[{Loshchilov and Hutter(2018)}]{loshchilov2018decoupled}
Ilya Loshchilov and Frank Hutter. 2018.
\newblock Decoupled weight decay regularization.
\newblock In \emph{International Conference on Learning Representations}.

\bibitem[{Ma et~al.(2019{\natexlab{a}})Ma, Lu, Wu, AlRegib, Kira, Socher, and
  Xiong}]{ma2019selfmonitoring}
Chih-Yao Ma, Jiasen Lu, Zuxuan Wu, Ghassan AlRegib, Zsolt Kira, Richard Socher,
  and Caiming Xiong. 2019{\natexlab{a}}.
\newblock \href {https://arxiv.org/abs/1901.03035} {Self-monitoring navigation
  agent via auxiliary progress estimation}.
\newblock In \emph{Proceedings of the International Conference on Learning
  Representations (ICLR)}.

\bibitem[{Ma et~al.(2019{\natexlab{b}})Ma, Wu, AlRegib, Xiong, and
  Kira}]{ma2019regretful}
Chih-Yao Ma, Zuxuan Wu, Ghassan AlRegib, Caiming Xiong, and Zsolt Kira.
  2019{\natexlab{b}}.
\newblock The regretful agent: Heuristic-aided navigation through progress
  estimation.
\newblock In \emph{Proceedings of the IEEE/CVF Conference on Computer Vision
  and Pattern Recognition}, pages 6732--6740.

\bibitem[{Mao et~al.(2016)Mao, Huang, Toshev, Camburu, Yuille, and
  Murphy}]{mao2016generation}
Junhua Mao, Jonathan Huang, Alexander Toshev, Oana Camburu, Alan~L Yuille, and
  Kevin Murphy. 2016.
\newblock Generation and comprehension of unambiguous object descriptions.
\newblock In \emph{Proceedings of the IEEE conference on computer vision and
  pattern recognition}, pages 11--20.

\bibitem[{Min et~al.(2021)Min, Chaplot, Ravikumar, Bisk, and
  Salakhutdinov}]{min2021film}
So~Yeon Min, Devendra~Singh Chaplot, Pradeep~Kumar Ravikumar, Yonatan Bisk, and
  Ruslan Salakhutdinov. 2021.
\newblock Film: Following instructions in language with modular methods.
\newblock In \emph{International Conference on Learning Representations}.

\bibitem[{Mnih et~al.(2016)Mnih, Badia, Mirza, Graves, Lillicrap, Harley,
  Silver, and Kavukcuoglu}]{mnih2016asynchronous}
Volodymyr Mnih, Adria~Puigdomenech Badia, Mehdi Mirza, Alex Graves, Timothy
  Lillicrap, Tim Harley, David Silver, and Koray Kavukcuoglu. 2016.
\newblock Asynchronous methods for deep reinforcement learning.
\newblock In \emph{International conference on machine learning}, pages
  1928--1937. PMLR.

\bibitem[{Moudgil et~al.(2021)Moudgil, Majumdar, Agrawal, Lee, and
  Batra}]{moudgil2021soat}
Abhinav Moudgil, Arjun Majumdar, Harsh Agrawal, Stefan Lee, and Dhruv Batra.
  2021.
\newblock Soat: A scene-and object-aware transformer for vision-and-language
  navigation.
\newblock \emph{Advances in Neural Information Processing Systems},
  34:7357--7367.

\bibitem[{Nguyen and Daum{\'e}~III(2019)}]{nguyen2019help}
Khanh Nguyen and Hal Daum{\'e}~III. 2019.
\newblock Help, anna! visual navigation with natural multimodal assistance via
  retrospective curiosity-encouraging imitation learning.
\newblock In \emph{Proceedings of the 2019 Conference on Empirical Methods in
  Natural Language Processing and the 9th International Joint Conference on
  Natural Language Processing (EMNLP-IJCNLP)}, pages 684--695.

\bibitem[{Qi et~al.(2020)Qi, Wu, Anderson, Wang, Wang, Shen, and
  Hengel}]{qi2020reverie}
Yuankai Qi, Qi~Wu, Peter Anderson, Xin Wang, William~Yang Wang, Chunhua Shen,
  and Anton van~den Hengel. 2020.
\newblock Reverie: Remote embodied visual referring expression in real indoor
  environments.
\newblock In \emph{Proceedings of the IEEE/CVF Conference on Computer Vision
  and Pattern Recognition}, pages 9982--9991.

\bibitem[{Qiao et~al.(2022)Qiao, Qi, Hong, Yu, Wang, and Wu}]{qiao2022hop}
Yanyuan Qiao, Yuankai Qi, Yicong Hong, Zheng Yu, Peng Wang, and Qi~Wu. 2022.
\newblock Hop: History-and-order aware pre-training for vision-and-language
  navigation.
\newblock In \emph{Proceedings of the IEEE/CVF Conference on Computer Vision
  and Pattern Recognition}, pages 15418--15427.

\bibitem[{Shah et~al.(2019)Shah, Chen, Rohrbach, and Parikh}]{shah2019cycle}
Meet Shah, Xinlei Chen, Marcus Rohrbach, and Devi Parikh. 2019.
\newblock Cycle-consistency for robust visual question answering.
\newblock In \emph{Proceedings of the IEEE/CVF Conference on Computer Vision
  and Pattern Recognition}, pages 6649--6658.

\bibitem[{Shridhar et~al.(2020)Shridhar, Thomason, Gordon, Bisk, Han, Mottaghi,
  Zettlemoyer, and Fox}]{shridhar2020alfred}
Mohit Shridhar, Jesse Thomason, Daniel Gordon, Yonatan Bisk, Winson Han,
  Roozbeh Mottaghi, Luke Zettlemoyer, and Dieter Fox. 2020.
\newblock Alfred: A benchmark for interpreting grounded instructions for
  everyday tasks.
\newblock In \emph{Proceedings of the IEEE/CVF conference on computer vision
  and pattern recognition}, pages 10740--10749.

\bibitem[{Tan et~al.(2019)Tan, Yu, and Bansal}]{tan2019learning}
Hao Tan, Licheng Yu, and Mohit Bansal. 2019.
\newblock Learning to navigate unseen environments: Back translation with
  environmental dropout.
\newblock In \emph{Proceedings of the 2019 Conference of the North American
  Chapter of the Association for Computational Linguistics: Human Language
  Technologies, Volume 1 (Long and Short Papers)}, pages 2610--2621.

\bibitem[{Wang et~al.(2021)Wang, Wang, Liang, Xiong, and
  Shen}]{wang2021structured}
Hanqing Wang, Wenguan Wang, Wei Liang, Caiming Xiong, and Jianbing Shen. 2021.
\newblock Structured scene memory for vision-language navigation.
\newblock In \emph{Proceedings of the IEEE/CVF Conference on Computer Vision
  and Pattern Recognition}, pages 8455--8464.

\bibitem[{Wang et~al.(2020)Wang, Wang, Shu, Liang, and Shen}]{wang2020active}
Hanqing Wang, Wenguan Wang, Tianmin Shu, Wei Liang, and Jianbing Shen. 2020.
\newblock Active visual information gathering for vision-language navigation.
\newblock In \emph{European Conference on Computer Vision}, pages 307--322.
  Springer.

\bibitem[{Wang et~al.(2019)Wang, Huang, Celikyilmaz, Gao, Shen, Wang, Wang, and
  Zhang}]{wang2019reinforced}
Xin Wang, Qiuyuan Huang, Asli Celikyilmaz, Jianfeng Gao, Dinghan Shen,
  Yuan-Fang Wang, William~Yang Wang, and Lei Zhang. 2019.
\newblock Reinforced cross-modal matching and self-supervised imitation
  learning for vision-language navigation.
\newblock In \emph{Proceedings of the IEEE/CVF Conference on Computer Vision
  and Pattern Recognition}, pages 6629--6638.

\bibitem[{Yu et~al.(2016)Yu, Poirson, Yang, Berg, and Berg}]{yu2016modeling}
Licheng Yu, Patrick Poirson, Shan Yang, Alexander~C Berg, and Tamara~L Berg.
  2016.
\newblock Modeling context in referring expressions.
\newblock In \emph{European Conference on Computer Vision}, pages 69--85.
  Springer.

\bibitem[{Zhu et~al.(2021{\natexlab{a}})Zhu, Liang, Zhu, Yu, Chang, and
  Liang}]{zhu2021soon}
Fengda Zhu, Xiwen Liang, Yi~Zhu, Qizhi Yu, Xiaojun Chang, and Xiaodan Liang.
  2021{\natexlab{a}}.
\newblock Soon: Scenario oriented object navigation with graph-based
  exploration.
\newblock In \emph{Proceedings of the IEEE/CVF Conference on Computer Vision
  and Pattern Recognition}, pages 12689--12699.

\bibitem[{Zhu et~al.(2020{\natexlab{a}})Zhu, Hu, Chen, Deng, Jain, Ie, and
  Sha}]{zhu2020babywalk}
Wang Zhu, Hexiang Hu, Jiacheng Chen, Zhiwei Deng, Vihan Jain, Eugene Ie, and
  Fei Sha. 2020{\natexlab{a}}.
\newblock Babywalk: Going farther in vision-and-language navigation by taking
  baby steps.
\newblock In \emph{Proceedings of the 58th Annual Meeting of the Association
  for Computational Linguistics}, pages 2539--2556.

\bibitem[{Zhu et~al.(2022)Zhu, Qi, Narayana, Sone, Basu, Wang, Wu, Eckstein,
  and Wang}]{zhu2022diagnosing}
Wanrong Zhu, Yuankai Qi, Pradyumna Narayana, Kazoo Sone, Sugato Basu, Xin~Eric
  Wang, Qi~Wu, Miguel Eckstein, and William~Yang Wang. 2022.
\newblock Diagnosing vision-and-language navigation: What really matters.
\newblock \emph{NAACL 2022}.

\bibitem[{Zhu et~al.(2021{\natexlab{b}})Zhu, Wang, Fu, Yan, Narayana, Sone,
  Basu, and Wang}]{zhu2021multimodal}
Wanrong Zhu, Xin Wang, Tsu-Jui Fu, An~Yan, Pradyumna Narayana, Kazoo Sone,
  Sugato Basu, and William~Yang Wang. 2021{\natexlab{b}}.
\newblock Multimodal text style transfer for outdoor vision-and-language
  navigation.
\newblock In \emph{Proceedings of the 16th Conference of the European Chapter
  of the Association for Computational Linguistics: Main Volume}, pages
  1207--1221.

\bibitem[{Zhu et~al.(2020{\natexlab{b}})Zhu, Wang, Narayana, Sone, Basu, and
  Wang}]{zhu2020towards}
Wanrong Zhu, Xin Wang, Pradyumna Narayana, Kazoo Sone, Sugato Basu, and
  William~Yang Wang. 2020{\natexlab{b}}.
\newblock Towards understanding sample variance in visually grounded language
  generation: Evaluations and observations.
\newblock In \emph{Proceedings of the 2020 Conference on Empirical Methods in
  Natural Language Processing (EMNLP)}, pages 8806--8811.

\end{thebibliography}
\bibliographystyle{acl_natbib}

\clearpage
\appendix

\section{ULN Datasets}\label{app:dataset}

\subsection{Data Collection}\label{app:data_collection}

\begin{table}[t]
  \begin{center}
  \resizebox{1.0\linewidth}{!}{
  \begin{tabular}{l|c|c|c|c}
    \hline \hline
    \multicolumn{1}{c|}{\multirow{2}{*}{Number of:}} &\multicolumn{1}{c|}{R2R} & \multicolumn{3}{c}{R2R-ULN} \\
    \cline{2-5} &
    \multicolumn{1}{c|}{L0} & \multicolumn{1}{c|}{L1} & 
    \multicolumn{1}{c|}{L2} & \multicolumn{1}{c}{L3} \\
    \hline \hline
    Instructions       & 1622   & 3282   & 3282   & 3282   \\
    
    Paths              & 917    & 917    & 917    & 917    \\
    
    Tokens             & 38.6   & 27.3   & 18.9   & 8.7    \\
    
    Direction Tokens   & 2.5    & 1.1    & 0.7    & 0.2    \\
    
    Object Tokens      & 8.5    & 6.6    & 4.6    & 2.6    \\
    \hline \hline
  \end{tabular}}
\end{center}
\caption{R2R validation v.s. our R2R-ULN validation. R2R-ULN preserves most of the paths from R2R. It is large-scale, human-centric, and strictly aligns all levels of underspecification.}
\label{tab:stats}
\end{table}

\begin{figure*}[t]
    \centering
    \includegraphics[width=0.9\linewidth]{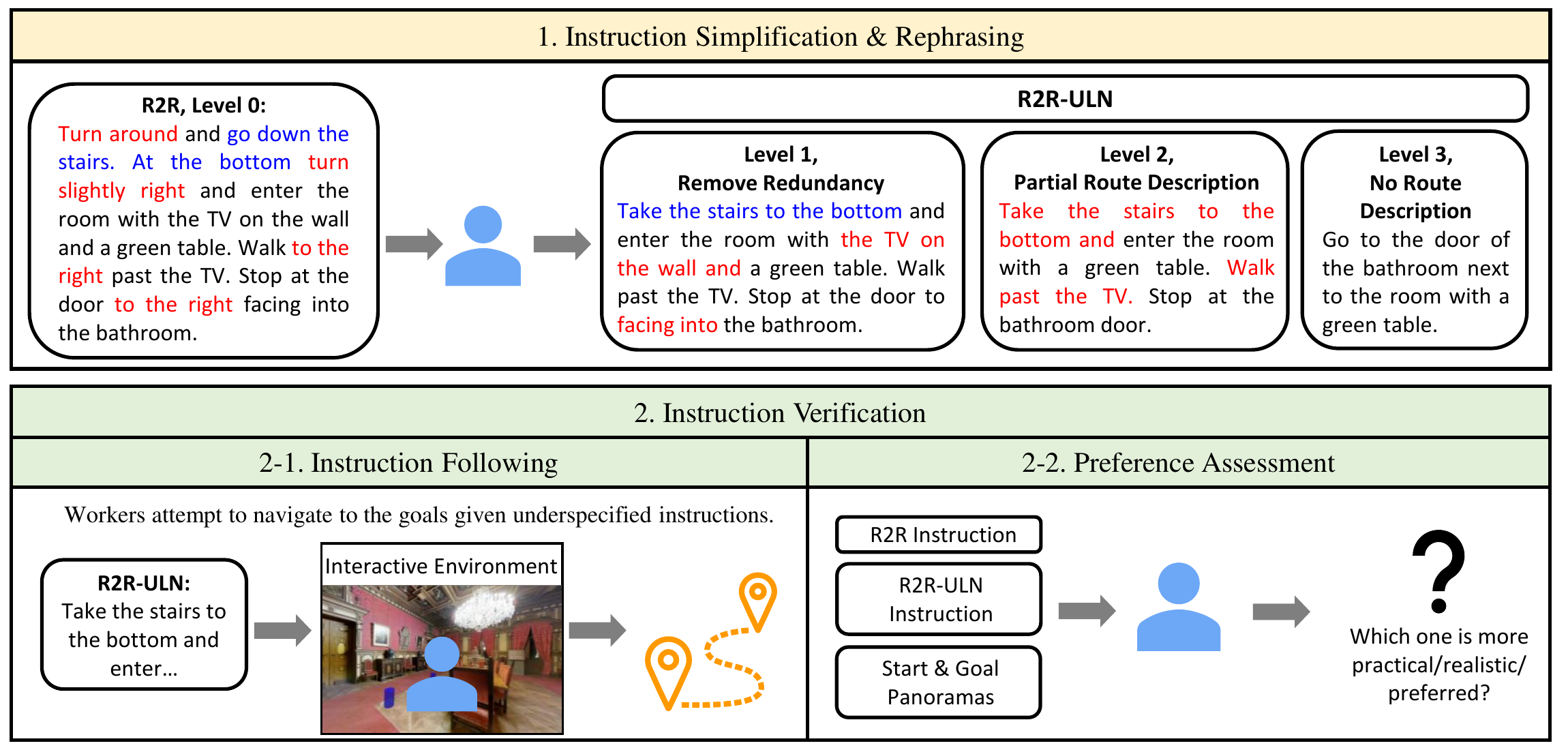}
    \caption{Dataset collection and verification. Text in \textcolor{red}{red} is removed in the next level and text in \textcolor{blue}{blue} is paraphrased in the next level.}
    \label{fig:collection}
\end{figure*}

Denote the original R2R instructions as Level 0 ($L_0$). We define Level 1 ($L_1$) instructions as ones where the redundant part is removed from $L_1$. Level 2~($L_2$) instructions are partial route description with some sub-instructions removed from $L_1$. Level 3~($L_3$) instructions directly refer to the goal destination without any intermediate route information.

\paragraph{Level 1:} 
We form the data collection stage as a sentence simplification task and ask human workers to progressively omit details from the instruction. For the first step, workers remove redundant parts from $L_0$ to obtain $L_1$. Redundancy includes but is not limited to repetition, excessive details, and directional phrases. To make it more human-centric, we allow the workers to determine the degree of redundancy. For example, some may rewrite ``turn 180 degrees'' as ``turn around'', while others may delete the whole phrase assuming that ``turn around'' is still redundant. 

\begin{table}[t]
  \begin{center}
  \resizebox{\linewidth}{!}{
  \begin{tabular}{l|l}
    \hline
    \multicolumn{1}{c|}{\textbf{Level}} & \multicolumn{1}{c}{\textbf{Instructions}} \\
    \hline \hline
    $L_0$ & \makecell[l]{\textcolor{red}{Go straight then slightly right to continue going straight.}\\ 
                        Exit the room then \textcolor{red}{turn left} and go into the room next door. \\ \textcolor{red}{Turn right} and go past the wall with the holes in it. Wait near \\ the lockers.} \\
    \hline
    $L_1$ & \makecell[l]{Exit the room and go into the room next door. \textcolor{red}{Go past} \\                                \textcolor{red}{the wall with holes in it.} Wait near the lockers.}\\
    \hline
    $L_2$ & \makecell[l]{\textcolor{red}{Exit the room} and go into the room next door. Wait near \\ 
                        the lockers.}\\
    \hline
    $L_3$ & \makecell[l]{Go into the room next door and wait near the lockers.}\\
    \midrule
    \midrule
    $L_0$ & \makecell[l]{\textcolor{red}{Turn around}, go through the kitchen and up the stairs \\                                \textcolor{red}{to the right.} \textcolor{red}{When at the top of the steps, turn to the left} \\ 
                        and go down the hallway. Stop in front of the painting \\ 
                        on the right wall. } \\
    \hline
    $L_1$ & \makecell[l]{\textcolor{red}{Go through the kitchen} and go up the stairs. Go down \\
                        the hallway. Stop in front of the painting on the wall.}\\
    \hline
    $L_2$ & \makecell[l]{\textcolor{red}{Go up the stairs} \textcolor{blue}{and go down the hallway.}\\                           \textcolor{blue}{Stop in front of the painting on the wall.}}\\
    \hline
    $L_3$ & \makecell[l]{Stop in front of the painting on the wall down the hallway.}\\
    \midrule
    \midrule
    $L_0$ & \makecell[l]{\textcolor{red}{Turn} and walk towards the open brown wooden door that \\ 
                        leads to an office with a large desk. Exit the room \textcolor{red}{through} \\ \textcolor{red}{the door}. Walk around \textcolor{red}{the left side of} the table and go \\ 
                        through the double open doors that lead to a hallway. \\ 
                        Walk out into the hallway until you reach the first door \\ 
                        \textcolor{red}{on the right}. \textcolor{red}{Turn tight} and take two steps into the room,\\ 
                        stopping in the doorway to the room next to the carpet.} \\
    \hline
    $L_1$ & \makecell[l]{\textcolor{blue}{Walk towards the open brown wooden door} that leads to \\ 
                        an office with a large desk. \textcolor{red}{Exit the room.} Walk around the table \\
                        and \textcolor{red}{go through the double open doors that leads to a hallway.}  \\ Walk out into the hallway \textcolor{red}{until you reach} \textcolor{red}{the first door.} \\ 
                        \textcolor{red}{Take two steps into the room}, stopping in the doorway to the room\\
                        next to the carpet. }\\
    \hline
    $L_2$ & \makecell[l]{\textcolor{red}{Take the open brown wooden door that leads into the office.}  \\ 
                        \textcolor{red}{Walk around the table} and \textcolor{blue}{go into the hallway.} \\ Stop in the doorway of the room next to the carpet.}\\
    \hline
    $L_3$ & \makecell[l]{Go to the doorway of the room next to the carpet in the hallway.}\\
    \hline
    \hline
  \end{tabular}
  }
\end{center}
\caption{More instruction examples from the R2R-ULN validation set. Words marked in red are removed, and words marked in blue are rephrased in the next level.}
\label{tab:uln_showcase_supp}
\end{table}

\paragraph{Level 2:} 
Then, one or two phrases containing objects are removed from $L_1$, resulting in partial route descriptions~($L_2$). $L_2$ assumes that humans tend to ignore intermediate references when the route is partially visible, or they are familiar with the environment.

\paragraph{Level 3:} Finally, the third step requires workers to write one sentence directly referring to the goals by combining $L_0$ and providing information like region label and floor level~($L_3$). $L_3$ resembles instructions in REVERIE~\cite{qi2020reverie} but is restricted to pure navigational instructions while REVERIE commands the agent to interact with target objects. As a result, we obtain one triplet ($L_1$, $L_2$, $L_3$) per $L_0$ instruction per annotator. 

Such progressive simplification design enables us to control the \textbf{inter-level alignment} as workers inject no external objects/directions or substitute existing ones with external ones for each $L_0$. It also preserves \textbf{intra-level sample variance} since workers simplify the sentences based on subjective judgments on the degree of redundancy, yet circumscribed by the definition of levels.

\paragraph{Instruction Following} We ask workers to reach the goal by following instructions and operating in an interactive WebGL environment. We randomly sample 250 triplets plus $L_0$ instructions from the validation set. We follow a similar setup as in \cite{anderson2018vision}. As is shown in Table~\ref{tab:human_verify}, workers can still achieve over 80\% success rate~(SR), the most suitable metric for ULN, on $L_1$ and $L_2$ and maintain a high-quality performance on $L_3$. Hence, ULN is a feasible setting where agents should actively make more explorations as the instructions become less specific.

\paragraph{Preference Assessment} As is shown in the bottom right of Fig.~\ref{fig:collection}, we investigate human preference on whether the full instructions or our collected ones are more practical and realistic. We sample 600 $L_1$-$L_3$ instructions and form each one with its corresponding $L_0$ instruction as a pair (Lx, $L_0$), x $\in [1,3]$. Given (Lx, $L_0$) and starting and goal viewpoint panoramas, workers are asked ``Q1. Which one is a more practical expression in daily life?'' and ``Q2. Which one do you prefer to speak to command the robot, considering applicability and efficiency?''. As shown in Table~\ref{tab:human_verify}, the results demonstrate an increasing trend in choosing shorter and less specific instructions to reflect more practical expressions and benefit applicability and efficiency.

\subsection{Dataset Statistics}\label{app:data_stats}
As illustrated in Table~\ref{tab:stats}, R2R-ULN preserves most of the trajectories from the R2R validation set. It addresses the shortage of underspecified instructions, which is essential for evaluating the generalization of embodied agents to various language expressions. Instructions of different lengths are better aligned, making it a better reference to investigate the correlation between instruction length and agent performance.

\section{Supplementary Experiments}\label{sec:supp_exp}

\subsection{Identifying GSS}
\begin{table}[t]
  \begin{center}
  \resizebox{0.9\linewidth}{!}{
  \begin{tabular}{l c cccc}
    \hline \hline
    \multicolumn{1}{c}{\multirow{3}{*}{Replaced part}} & & \multicolumn{4}{c}{R2R-ULN Val-Unseen}\\
    & & \multicolumn{4}{c}{Level 3 ($L_3$)}\\
    \cline{3-6} & &
    \multicolumn{1}{c}{TL} & \multicolumn{1}{c}{NE$\downarrow$} & \multicolumn{1}{c}{SR$\uparrow$} & \multicolumn{1}{c}{SPL$\uparrow$}\\
    \hline
    \textit{VLN$\protect\CircleArrowright$BERT} \\
    low-level && 13.20 & 7.41 & 32.71 & 29.07\\
    $f_{\text{text}}$ && 14.23 & 7.18 & 36.64 & 31.25\\
    $f_{\text{text}}, f_{\text{emb}}$  && 14.26 & 7.26 & 36.37 & 31.18\\
    $f_{\text{img}}$ && 12.31 & 7.38 & 33.07 & 29.27\\
    $f_{\text{cm}}$ && 13.50 & 7.15 & 34.94 & 31.22\\
    high-level && 13.31 & 6.96 & 36.45 & 31.99\\
    \hline
    \textit{HAMT} \\
    low-level && 13.26 & 7.18 & 35.12 & 31.13 \\
    $f_{\text{text}}$ && 13.26 & 7.18 & 35.12 & 31.13\\
    $f_{\text{hist}}$ && 13.26 & 7.18 & 35.12 & 31.13\\
    $f_{\text{img}}$  && 13.22 & 7.28 & 34.23 & 30.38\\
    $f_{\text{cm}}$  &&  13.26 & 6.75 & 38.74 & 34.86\\
    high-level && 13.34 & 6.95 & 37.18 & 33.40 \\
    \hline \hline
  \end{tabular}
  }
\end{center}
\caption{Low-level baseline performance after replacing sub-network from another high-level agent with the same architecture. The low-level agent is one trained with full R2R instructions. The high-level agent is one trained with last sentence of R2R instructions.}
\label{tab:gss_appendix}
\end{table}

Intuitively, this sub-network incorporates into the cross-modal transformer. The conjecture is that the cross-attention layers shift the vision-to-text attention from the first token to the end for low-level instructions. On the other hand, for high-level instructions, the visual features may persistently attend to the goal object tokens. To verify this intuition, 

We show three additional experimental results here for supplementary. First, as is shown in Table \ref{tab:gss_appendix}, we identify the critical sub-network by replacing part of a low-level agent with that from a high-level agent and observe the performance change. For VLN$\protect\CircleArrowright$BERT, the most crucial sub-network that impacts its navigation mode is the text encoder, as replacing it achieves the highest performance. Note that replacing the embedding layer downgrades the performance. We hypothesize that the high-level agent is trained with a smaller vocabulary due to shorter instructions. As for HAMT, the critical sub-network is the cross-modal encoder, which aligns with our hypothesis.

\subsection{Ablation Studies} \label{app:ablation}

\begin{table}[t]
  \begin{center}
  \resizebox{1.0\linewidth}{!}{
  \begin{tabular}{l|cc|cc}
    \hline \hline
    \multicolumn{1}{c}{\multirow{2}{*}{Methods}} &\multicolumn{2}{|c|}{R2R Val-Seen} & \multicolumn{2}{c}{R2R Val-Unseen} \\
    \cline{2-5} &
    \multicolumn{1}{c}{SR$\uparrow$} & \multicolumn{1}{c|}{SPL$\uparrow$} & 
    \multicolumn{1}{c}{SR$\uparrow$} & \multicolumn{1}{c}{SPL$\uparrow$} \\
    \hline \hline
    EnvDrop &62.3&59.4&50.0&46.8\\
    PREVALENT &69.5&65.4&58.3&53.6\\
    \hline
    FAST-Short &-&-&55.8&43.3 \\
    SSM &71.1&61.8&62.4&45.3\\
    \hline
    VLN$\protect\CircleArrowright$BERT & \textbf{74.2} & \textbf{69.8} & 61.4 & \underline{55.6} \\
    Ours~(w/o E2E)         & \underline{73.9} & \underline{69.3} & \underline{61.6} & \textbf{55.8} \\
    Ours~(w/ E2E)         & 73.0 & 57.3 & \textbf{62.8} & \underline{45.0} \\
    \midrule
    HAMT           & 75.6 & 72.2 & 66.2 & 61.5 \\
    Ours~(w/o E2E) & 73.6 & 70.2 & 65.6 & 60.7 \\
    Ours~(w/ E2E)  & 73.3 & 51.9 & 64.4 & 41.3 \\
    \hline \hline
  \end{tabular}}
\end{center}
\caption{Performance on R2R validation set in the single-run setting. We disable E2E since ULN does not encourage exploration with $L_0$ instructions.}
\label{tab:r2r_result_appendix}
\end{table}
\paragraph{R2R Evaluation} We also show full performance results on R2R in Table \ref{tab:r2r_result_appendix}. Our framework maintains a comparable performance on R2R for VLN$\protect\CircleArrowright$BERT and even slightly improves the SR with E2E. However, as for HAMT, our framework downgrades the SR by 0.6\%. This is due to the imperfect classifier that misclassifies some short R2R instructions as high-level instructions. We empirically find that the HAMT is more sensitive to classifier error, indicating better robustness of HAMT in handling short R2R instructions. It spares the need to run the high-level sub-network to deal with those exceptions in R2R.

\begin{table}[t]
  \begin{center}
  \resizebox{1.0\linewidth}{!}{
  \begin{tabular}{l|c|cc}
    \hline \hline
    \multicolumn{1}{c|}{\multirow{3}{*}{\makecell[c]{Training\\ Set}}} & \multicolumn{1}{c|}{\multirow{3}{*}{\makecell[c]{\# of training \\instr.}}} & \multicolumn{2}{c}{R2R-ULN Val-Unseen} \\
    \cline{3-4} &
    & \multicolumn{2}{c}{L3} \\
    \cline{3-4} &
    & \multicolumn{1}{c}{SR$\uparrow$} & \multicolumn{1}{c}{SPL$\uparrow$} \\
    \hline \hline
    R2R-Last+Speaker & 24505
        & \textbf{34.8} & \textbf{31.5} \\
    R2R-Last+REVERIE & 24505
        & 32.3 & 28.3 \\
    R2R-Last+SOON & 16718
        & 34.0 & 29.3 \\
    \hline \hline
  \end{tabular}}
\end{center}
\caption{Agents trained with a combination of R2R-Last and one of the existing high-level datasets, REVERIE or SOON.}
\label{tab:ablation_soon_reverie}
\end{table}
\paragraph{Combining Multiple Datasets} Since there are existing high-level datasets like REVERIE \cite{qi2020reverie} and SOON \cite{zhu2021soon}, there is also a natural motivation to combine these datasets during training since these instructions can be seen as goal instructions (similar to $L_3$). We trained three agents with a combination of R2R-Last and one of the high-level datasets. For comparison, we also train an agent with R2R-Last, and the same amount of speaker-augmented instructions. Note that the speaker is trained with last sentences as well, so it is a goal instruction speaker. As is shown in Table \ref{tab:ablation_soon_reverie}, existing datasets have limited benefit on $L_3$ evaluation. REVERIE and SOON instructions contain both navigation and localization orders, making these datasets noisier for pure navigation training and evaluation. We leave it as future work to consider combining VLN datasets of different settings for one unified, general navigational agent.

\begin{table}[t]
  \begin{center}
  \resizebox{\linewidth}{!}{
  \begin{tabular}{cccc c ll ll ll}
    \hline \hline
    \multicolumn{4}{c}{Components} && \multicolumn{6}{c}{R2R-ULN Val-Unseen} \\
    \cline{1-4} \cline{6-11}
    \multicolumn{1}{c}{\multirow{2}{*}{Classify Instr.}} & \multicolumn{1}{c}{\multirow{2}{*}{GSS}} & \multicolumn{2}{c}{E2E} && \multicolumn{2}{c}{L1} & \multicolumn{2}{c}{L2} & \multicolumn{2}{c}{L3} \\
    \cline{3-4} \cline{6-11} 
     &  & \multicolumn{1}{c}{Lookahead} & \multicolumn{1}{c}{State Freeze} &&
    \multicolumn{1}{c}{SR$\uparrow$} & \multicolumn{1}{c}{SPL$\uparrow$} & 
    \multicolumn{1}{c}{SR$\uparrow$} & \multicolumn{1}{c}{SPL$\uparrow$} & 
    \multicolumn{1}{c}{SR$\uparrow$} & \multicolumn{1}{c}{SPL$\uparrow$} \\
    \hline
     & & & && 50.8 & \textbf{45.4} & 44.1 & 39.5 & 32.7 & 29.1 \\
     & \checkmark & & && 50.8 & \textbf{45.4} & 44.1 & 39.5 & 37.8 & \textbf{33.5} \\
     \checkmark & \checkmark & & && 51.0 & 45.6 & 45.1 & \textbf{40.4} & 37.8 & \textbf{33.5} \\
     & & \checkmark & && 51.8 & 37.1 & 46.7 & 31.9 & 34.5 & 21.3 \\
     & & \checkmark & \checkmark && \textbf{52.6} & 36.7 & 46.6 & 31.3 & 35.9 & 21.5 \\
     \checkmark & \checkmark & \checkmark & & & 52.0 & 37.3 & 46.8 & 32.4 & 37.8 & 24.6 \\
     \checkmark & \checkmark & \checkmark & \checkmark && \textbf{52.7} & 36.9 & \textbf{47.4} & 32.5 & \textbf{39.5} & 25.6 \\
    \hline \hline
  \end{tabular}
  }
\end{center}
\caption{Full ablation study with SPL reported.}
\label{tab:ablation_appendix}
\end{table}

\begin{figure}[t]
    \centering
    \includegraphics[width=\linewidth]{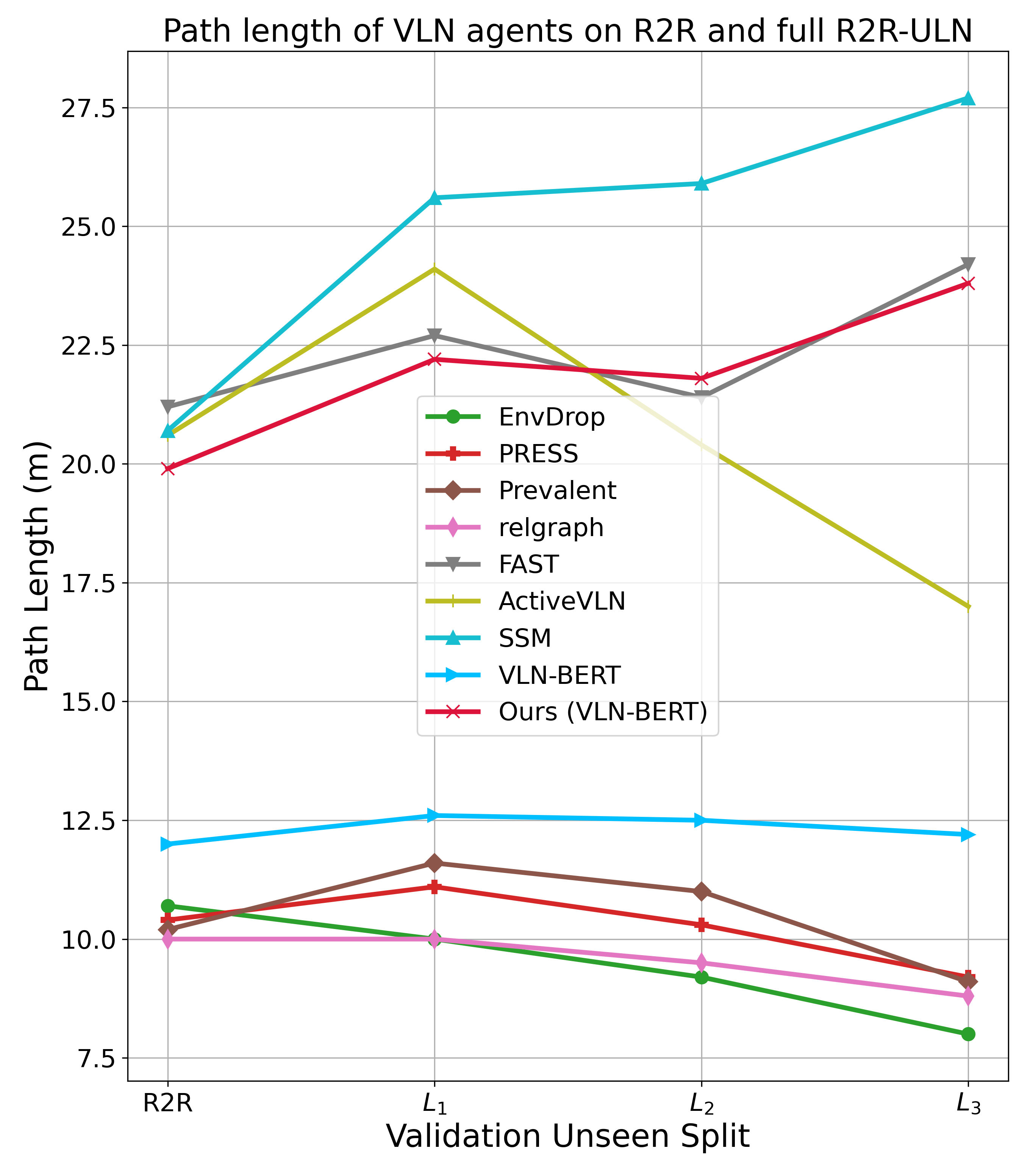}
    \caption{Path length of various agents on R2R and R2R-ULN.}
    \label{fig:all_agents_pl}
\end{figure}

\begin{figure}[t]
    \centering
    \includegraphics[width=\linewidth]{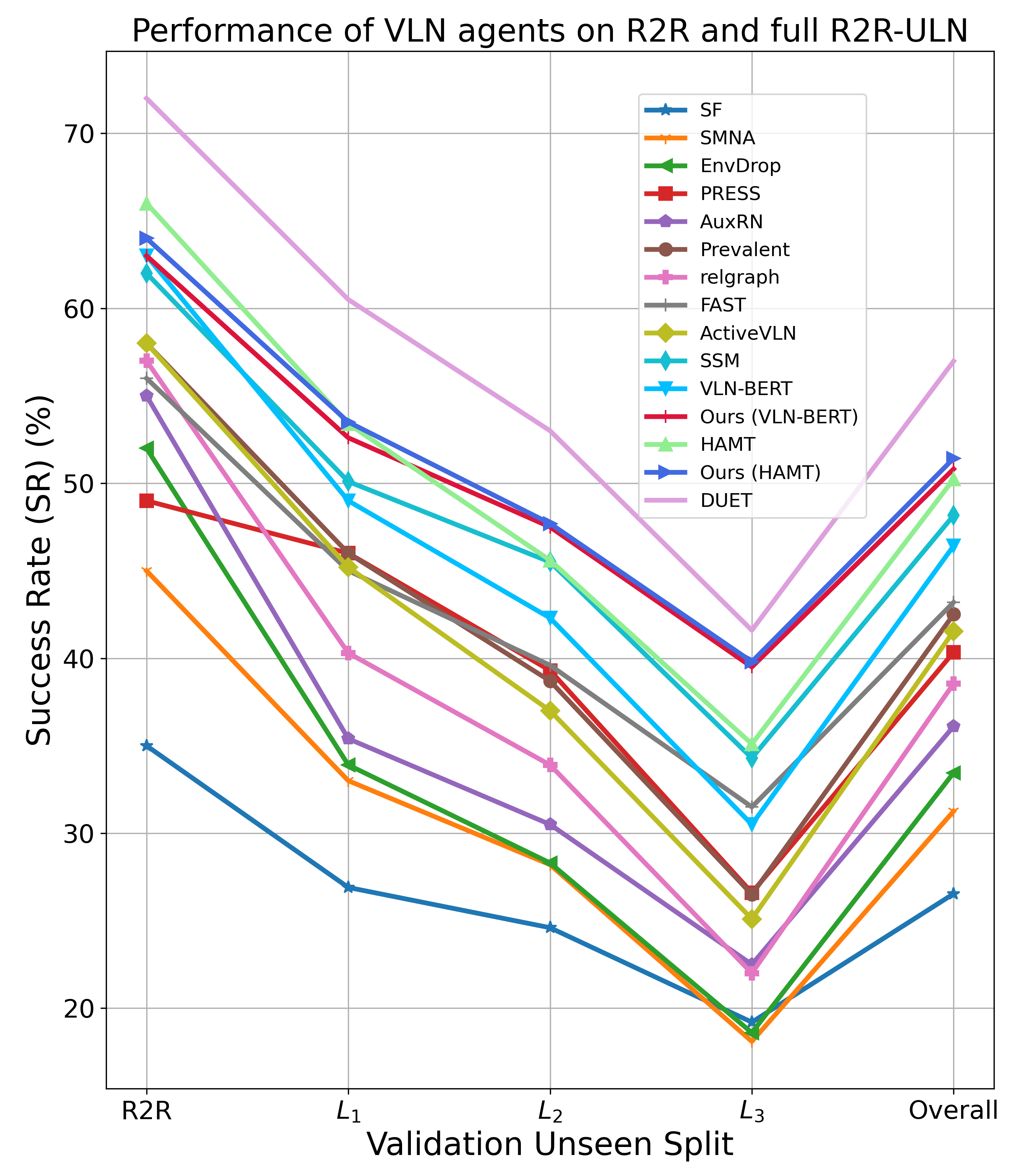}
    \caption{Evaluation results of many base agents on R2R and full R2R-ULN.}
    \label{fig:all_agents_sr}
\end{figure}

\paragraph{Path Length} As is shown in Fig. \ref{fig:all_agents_pl}, most greedy-decoding agents learn a spurious correlation between instruction length and trajectory length (TL), reflected as a decreasing trend in TL from $L_1$ to $L_3$. However, SOTA agents are more robust as they navigate slightly longer for L3. Exploration-based agents showed significantly longer TL and increasing trends, which is more robust (in SR percentage drop) and interpretable.

\paragraph{Component Analysis} Last but not least, we show the full ablation table with SPL compared to Table \ref{tab:ablation}. Our GSS-only framework achieves the highest SPL while adding the E2E module decreases the SPL value. This is intuitive as the purpose of E2E is to improve the success rate by making more exploration steps, and will result in a smaller SPL value.

\subsection{Trajectory Visualization}
\begin{figure*}[t]
    \centering
    \includegraphics[width=\textwidth]{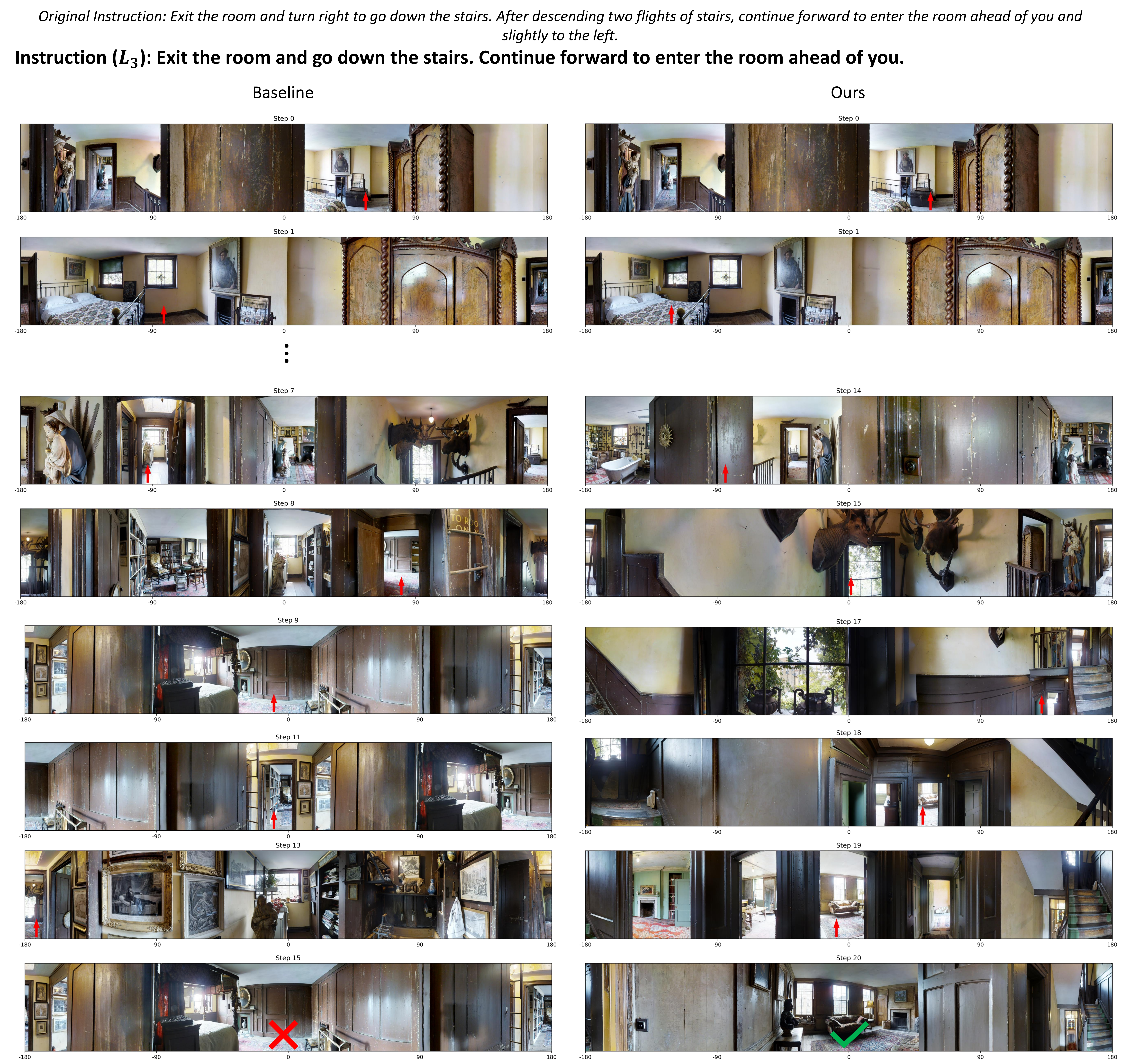}
    \caption{Visualization of trajectories following an $L_1$ instruction.}
    \label{fig:vis_l1}
\end{figure*}

\begin{figure*}[t]
    \centering
    \includegraphics[width=\textwidth]{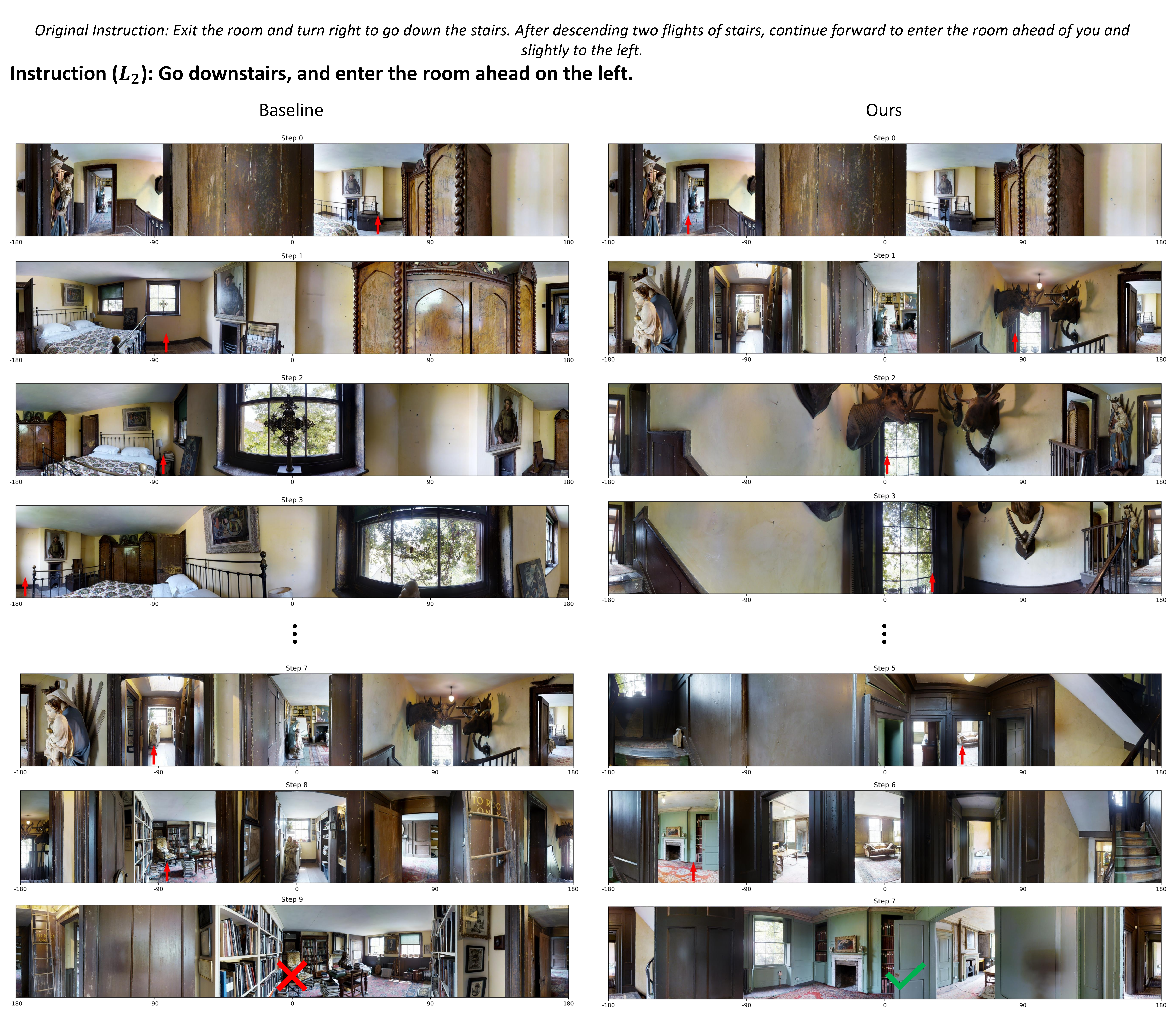}
    \caption{Visualization of trajectories following an $L_2$ instruction.}
    \label{fig:vis_l2}
\end{figure*}

We provide visualizations of a set of underspecified instruction ($L_1, L_2, L_3$) in add to Fig. \ref{fig:teaser}. As is shown in Fig. \ref{fig:vis_l1}, the base agent fails to go downstairs from the beginning and keeps moving around on the initial floor. Our agent also navigates on the initial floor for a long time. However, it starts moving downstairs from step 14 thanks to uncertainty estimation and active exploration at the previous step. Finally, it reaches the lounge downstairs. Fig. \ref{fig:vis_l2} shows the trajectories of an $L_2$ instruction. Similarly, the base agent keeps moving around on the initial floor and fails to go downstairs. However, our agent classifies the instruction as high-level and adopts the high-level GSS to guide the navigation decisions. Therefore, it reaches the goal with only seven steps. As for $L_3$, the base agent manages to go downstairs at step 7 but keeps moving downstairs and completely misses out on the lounge, resulting in a long path. In contrast, our agent with the high-level GSS reaches the target location with only seven steps. There is no exploration because no uncertain steps are identified.

\begin{figure*}[t]
    \centering
    \includegraphics[width=\textwidth]{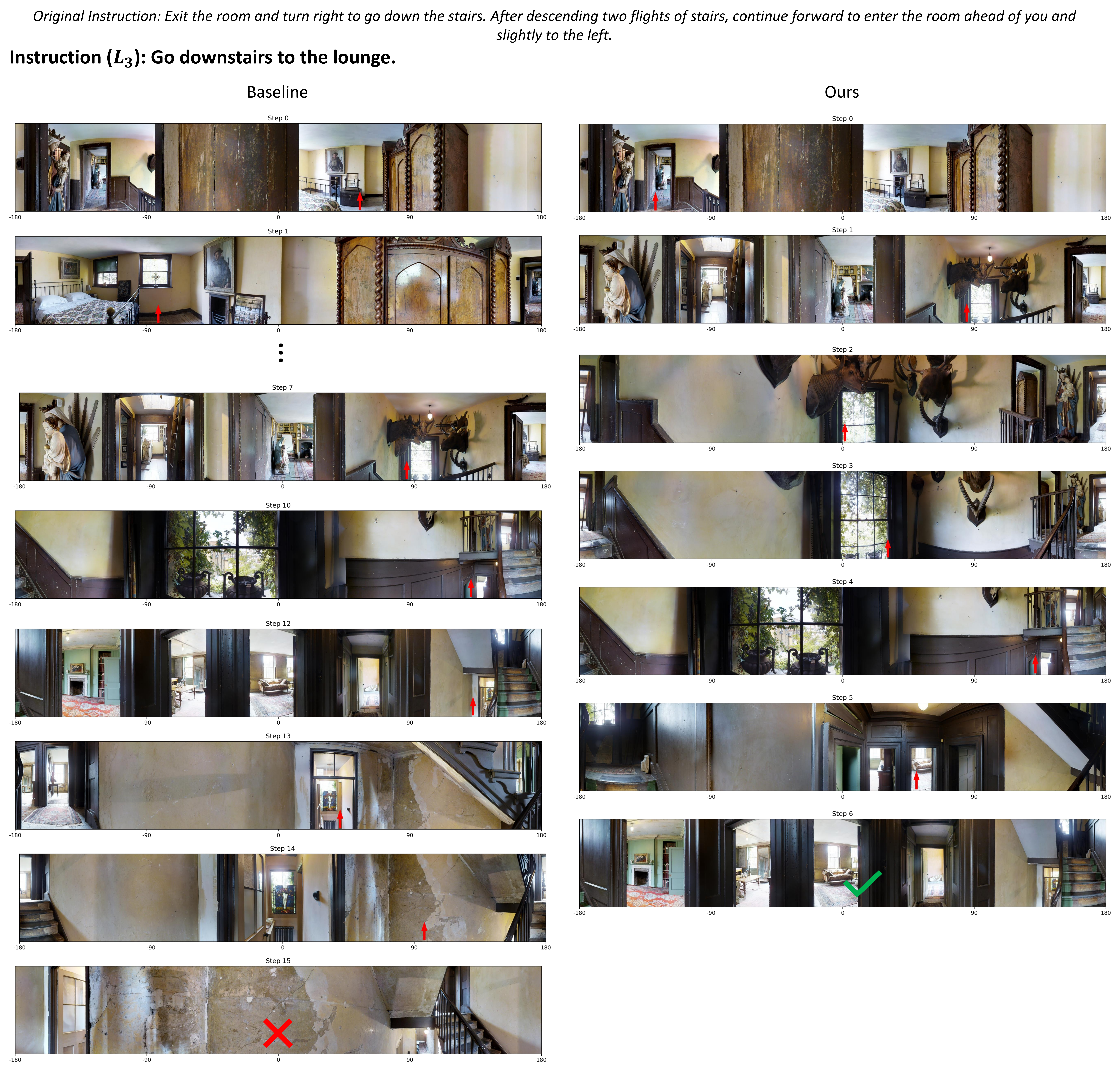}
    \caption{Visualization of trajectories following an $L_3$ instruction.}
    \label{fig:vis_l3}
\end{figure*}

\end{document}